\documentclass[letterpaper]{article} % DO NOT CHANGE THIS
\usepackage{aaai24}  % DO NOT CHANGE THIS
\usepackage{times}  % DO NOT CHANGE THIS
\usepackage{helvet}  % DO NOT CHANGE THIS
\usepackage{courier}  % DO NOT CHANGE THIS
\usepackage[hyphens]{url}  % DO NOT CHANGE THIS
\usepackage{graphicx} % DO NOT CHANGE THIS
\usepackage{natbib}  % DO NOT CHANGE THIS AND DO NOT ADD ANY OPTIONS TO IT
\usepackage{caption} % DO NOT CHANGE THIS AND DO NOT ADD ANY OPTIONS TO IT
\usepackage{algorithm}
\usepackage[noend]{algpseudocode}
\usepackage{tikz,pgfplots}
\usepackage{subfig}
\usepackage{newfloat}
\usepackage{listings}
\DeclareCaptionStyle{ruled}{labelfont=normalfont,labelsep=colon,strut=off} % DO NOT CHANGE THIS
\floatstyle{ruled}
\newfloat{listing}{tb}{lst}{}
\floatname{listing}{Listing}
\title{Fusing Conditional Submodular GAN and Programmatic Weak Supervision}
\author{
    Kumar Shubham, 
    Pranav Sastry,
    Prathosh AP
}
\affiliations{
    Indian Institute of Science,
    Bangalore, India \\
    shubhamkuma3@iisc.ac.in, pranavsastry@iisc.ac.in, prathosh@iisc.ac.in
}

\usepackage{bibentry}
\usepackage{xcolor}
\usepackage{tikz}
\usepackage{amsmath,amsfonts,bm}

\def\eqref#1{equation~\ref{#1}}
\def\1{\bm{1}}

\def\vone{{\bm{1}}}

\def\vl{{\bm{l}}}

\def\vs{{\bm{s}}}

\def\vv{{\bm{v}}}

\DeclareMathAlphabet{\mathsfit}{\encodingdefault}{\sfdefault}{m}{sl}
\SetMathAlphabet{\mathsfit}{bold}{\encodingdefault}{\sfdefault}{bx}{n}

\def\gA{{\mathcal{A}}}
\def\gB{{\mathcal{B}}}
\def\gC{{\mathcal{C}}}
\def\gD{{\mathcal{D}}}

\def\gF{{\mathcal{F}}}
\def\gG{{\mathcal{G}}}
\def\gH{{\mathcal{H}}}

\def\gK{{\mathcal{K}}}
\def\gL{{\mathcal{L}}}

\def\gP{{\mathcal{P}}}

\def\gR{{\mathcal{R}}}
\def\gS{{\mathcal{S}}}

\def\gV{{\mathcal{V}}}

\def\gY{{\mathcal{Y}}}

\newcommand{\E}{\mathbb{E}}

\DeclareMathOperator*{\argmax}{arg\,max}

\newtheorem{prop}{Proposition}
\newtheorem{definition}{Definition}
\begin{document}
\maketitle
\begin{abstract}
    Programmatic Weak Supervision (PWS) and generative models serve as crucial tools that enable researchers to maximize the utility of existing datasets without resorting to laborious data gathering and manual annotation processes. PWS uses various weak supervision techniques to estimate the underlying class labels of data, while generative models primarily concentrate on sampling from the underlying distribution of the given dataset. Although these methods have the potential to complement each other, they have mostly been studied independently.
 Recently, WSGAN proposed a mechanism to fuse these two models. Their approach utilizes the discrete latent factors of InfoGAN to train the label model and leverages the class-dependent information of the label model to generate images of specific classes. However, the disentangled latent factors learned by InfoGAN might not necessarily be class-specific and could potentially affect the label model's accuracy. Moreover, prediction made by the label model is often noisy in nature and can have a detrimental impact on the quality of images generated by GAN. In our work, we address these challenges by (i) implementing a noise-aware classifier using the pseudo labels generated by the label model (ii) utilizing the noise-aware classifier's prediction to train the label model and generate class-conditional images. Additionally, we also investigate the effect of training the classifier with a subset of the dataset within a defined uncertainty budget on pseudo labels. We accomplish this by formalizing the subset selection problem as a submodular maximization objective with a knapsack constraint on the entropy of pseudo labels. We conduct experiments on multiple datasets and demonstrate the efficacy of our methods on several tasks vis-a-vis the current state-of-the-art methods. Our implementation is available at \textit{https://github.com/kyrs/subpws-gan}
\end{abstract}
\section{Introduction}

The success of many deep learning methods is often credited to the availability of large amounts of labeled data ~\cite{imagenet,celeb-500k,hypersim-data,vggsound-data}. However, labeling necessitates manual annotation, which is a time-consuming and labor-intensive process.  Distant supervision or weak supervision methods address this issue by using an external knowledge base~\cite{distant_supervision_KB}, pre-trained models~\cite{bach2019snorkel}, and heuristics approaches~\cite{distant_supervision_heuristic} to generate inexpensive but potentially noisy class labels for downstream tasks. These methods use an alternative source of information to assign labels to the training data and thus avoid the need for extensive manual annotation. However, the pseudo labels generated by such methods are often noisy in nature and can have a detrimental impact on downstream performance.

Programmatic weak supervision~\cite{PWS_survey, PWS_dehghani,PWS_lang, PWS_ratner2016,PWS_ratner2020snorkel,PWS_riedel} addresses this issue by combining the prediction of multiple such noisy sources of labels often known as label functions (LFs) systematically. In PWS, a label model is trained to combine the prediction of these label functions to estimate the unobserved ground truth label for a given training sample. The generated pseudo label is then used for downstream training.  For a given set of label functions, the quality of the pseudo label is often determined by the performance of the label model.
%Generally, the label functions often have correlated and conflicting prediction and the key objective of the label model is to efficiently weight the predictions of these label function to generate accurate pseudo label. 
Various approaches have been suggested to effectively train the label model~\cite{PWS_survey} without knowing the underlying ground truth class labels. While approaches like~\cite{PWS_ratner2016, PWS_ratner2020snorkel,PWS_ratner2019_metal,PWS_fu_fastsquid} use only the prediction of label functions to estimate the pseudo label, encoder-based label models also consider data features to estimate pseudo labels~\cite{cauchy_end_to_end, gen_model_PWS_vice_versa}.

% These label functions often have correlated and conflicting prediction and the key objective of PWS is to train a label model that can combine the outputs from multiple label functions to generate a final pseudolabel for the complete dataset without utilizing any ground truth label information. This pseudolabel, along with the corresponding dataset, is then utilized for the downstream training task.

% The inter-dependency between the label functions can impact the quality of pseudolabels generated by the labelmodel. For example, a naive approach to generate pseudolabels is to take a majority vote of the prediction of label functions. However, in a setting where most of the label function have correlated output, this can lead to a biased prediction. Recent works address this by allowing users to manually incorporate the dependency structure in label model's prediction[REF] or learn these dependency and associated errors in unsupervised fashion. [REF] further propose to generate sample dependent pseudolabels using the data dependent feature in contrast with other methods that only relying on the prediction of the label functions.  

 Recently,~\citet{gen_model_PWS_vice_versa} proposed WSGAN, a novel training strategy to efficiently generate pseudo labels for a given image dataset. Their approach combines InfoGAN with an encoder-based label model~\cite{cauchy_end_to_end}. The key idea is to train both InfoGAN~\cite{infogan} and label model simultaneously and leverage the disentangled latent factors learned by InfoGAN to align the label model's predictions. 
 % The pseudo labels associated with each sample can further be utilized to enhance the class-specific features in the disentangled representation. 
 This training process offers several advantages. The synchronized training of the label model and InfoGAN allows the efficient utilization of the disentangled latent factors of InfoGAN for the training of the label model. Simultaneously, the pseudo labels generated by the label model facilitate the generation of class-conditioned images, which can benefit downstream tasks as an additional form of data augmentation. 
 
 Nevertheless, the implementation of such a training process presents several challenges. The disentangled latent representation of InfoGAN is learned by maximizing the mutual information between the discrete conditional input and the generated images. However, the representations (latent factors) learned from InfoGAN may not be class-specific. In such a scenario, aligning the prediction of the label model with such latent factors can hurt the accuracy of the label model. Furthermore, the predictions of the label models are often noisy in nature, and using the pseudo labels directly for GAN training can make the overall training process unstable, impacting the quality of images generated by InfoGAN.

In this study, we address these challenges by training a noise-aware classifier-guided conditional GAN~\cite{ACGAN} for the training of label model and synthetic image generation. In contrast to conventional conditional GANs that rely on authentic ground truth labels during training, our approach operates within a weak supervision framework, where the pseudo labels generated by the label model are used to train the classifier. To handle the noise associated with the pseudo labels, the classifier is trained using a noise-aware symmetric cross entropy loss~\cite{symmetric_cross_entropy} in an adaptive fashion~\cite{psuedolabel_gan_Morerio}. Within the current setup, as training proceeds, a new set of refined dataset and associated pseudo labels generated by the label model is used to train the classifier. 
% Recent studies have shown that in weak supervision setting, a subset of most representative samples with least entropy associated with pseudolabels generates better performance than complete dataset.  Taking inspiration from these works, we further investigate the impact of training a classifier under a noisy setting, with a subset of most representative and diverse example with least entropy associated with the pseudolabels. For this, we formulate  the data subset selection process as a submodular maximization problem under the knapsack constraint [REF] defined over the entropy of the pseudolabels predicted by the labelmodel. 

Further, a recent line of research by~\citet{subset_ws_neurips22} shows that, in weak supervision, a subset of the most representative training samples yields superior performance compared to using the entire dataset for downstream training of a classifier. Motivated by this,  we propose to utilize a subset of highly representative and diverse examples that exhibit minimal uncertainty with the pseudo labels. We accomplish this by incorporating a data subset selection process for weak supervision by formulating it as a submodular maximization problem~\cite{submodularity_AI_blimes,krause_submodular_maximization} with a knapsack constraint defined over the overall entropy of the pseudo labels of the selected samples. ~\cite{krause_submodular_maximization,submodular_knapsack_NAS}.  Our contributions are summarized below:

(1) We propose a novel technique to fuse a classifier-guided conditional GAN with an encoder-based label model. Within the framework of programmatic weak supervision (PWS), this helps in efficient training of the label model and generation of class conditional images, 
(2) We present a novel approach for subset selection to be used in tandem with PWS by identifying the most diverse and representative samples via a submodular maximization technique, aiding in the reduction of the uncertainty associated with the training dataset, and (3) We investigate the impact on the overall performance of the label model as well as on the quality of images using the proposed subset selection scheme and compare our method with the current state-of-the-art.

\section{Related Work}

\subsection{Programmatic Weak Supervision}
 Programmatic Weak Supervision~\cite{PWS_ratner2016} is an efficient technique that addresses the issue of lack of ground truth labels for a downstream task. 
 Within this framework, a subject matter expert uses different sources of noisy labeling schemes to annotate the unlabeled dataset. These sources often known as label functions (LFs) provides partial information about the true labels and exhibit superior performance than a random model. Generally, these label functions are often approximated using an external knowledge base~\cite{distant_supervision_KB}, pre-trained models~\cite{bach2019snorkel}, heuristics approaches~\cite{distant_supervision_heuristic}
and other similar techniques~\cite{PWS_survey}. A common practice in PWS is to define a threshold associated with each of these label functions so that an LF can abstain from making an uncertain prediction about any example. One of the main tasks in PWS is to train a label model that combines the prediction from different LFs to estimate the underlying ground truth labels. This is done by estimating the accuracy and dependency between LFs and then using it to weight their prediction to estimate the pseudo labels. Different methods have used different approximation techniques to make informed predictions about the pseudo labels. \citet{PWS_ratner2016} modeled the label model using a factor graph, \citet{PWS_dawid1979} used expectation maximization to estimate the pseudo labels, \citet{PWS_ratner2019_metal} used a Markov network and matrix completion techniques to recover the associated parameters, FlyingSquid~\cite{PWS_fu_fastsquid} used ising model to predict pseudo labels. While most of the methods make an estimate about the pseudo labels by only considering the prediction made by the label functions, \citet{cauchy_end_to_end} proposed to use a neural network, which implicitly captures these dependencies and use data-dependent features to estimate the accuracy of the label functions. WSGAN~\cite{gen_model_PWS_vice_versa} and WSVAE~\cite{PWS_VAE}  further built on top of the idea proposed by \citet{cauchy_end_to_end}, and use generative models such as InfoGAN and VAE to train the label model. While WSGAN proposed the given fusion for image data, WSVAE primarily focused on improving the label model's performance. To improve the performance of the downstream model, recent works~\cite{ NEURIPS2022_1343edb2, pmlr-v151-yu22c} have proposed to use noise-aware loss in PWS. However, these losses utilize the weights of label model to improve the downstream prediction task and, hence, are not suitable for the training of the label model.
 \\
 Label models are generally data agnostic and only depend on the prediction made by the label functions. However, the label function used to generate noisy labels can be different for different types of data modalities. For instance, \citet{varma2018snuba,fries2019weakly} proposed to generate pseudo labels using heuristic rules. Similarly, \citet{chen2019slice,hooper2020cut} define rules on top of the prediction made by a surrogate model. \citet{ joulin2016learning, wang2017chestx,irvin2019chexpert,boecking2020interactive,eyuboglu2021multi} used the information from other modalities of data like text to estimate the labels.\\ %Once these noisy labels have been generated, they can either be used directly for the training of the downstream task or a refined label generated by a label model using prediction from multiple sources can be used for the training.  

\subsection{Conditional GANs with noisy labels}
% The synthetic dataset generated by a class-conditional generative model, such as Generative Adversarial Network (GAN)~\cite{GAN}, is used in PWS where the labels are noisy. 
Synthetic data generated using class-conditional generative models, such as Generative Adversarial Networks (GANs)~\cite{GAN} is another approach to efficiently utilize the available dataset. However, in a weak supervision setting, such a model needs to be trained with noisy class labels.
In a noisy environment, several methods have been proposed to efficiently train conditional GAN~\cite{thekumparampil2018robustness,han2020robust,kaneko2019label,kaneko2021blur}. However, most of these methods use a predefined set of noisy class labels to train the GAN. Recently, \citet{psuedolabel_gan_Morerio} have shown that in a multi-domain setting, the performance of a classifier can further be improved by utilizing a conditional GAN~\cite{cGAN}. They proposed a unique training paradigm where the GAN and the classifier are jointly trained in an iterative fashion, in which the classifier provides refined labels for the GAN, and the GAN provides better samples to train the classifier. However, the training of such a model is limited only to a multi-domain setting, where accurate class labels for at least one domain are available.  

\subsection{Subset selection for weak supervision}

 The performance of a classifier in a weakly supervised setting is influenced by the data used for its training~\cite{subset_ws_neurips22,angelova2005pruning,maheshwari2020semi,mirzasoleiman2020coresets,karamanolakis-etal-2021-self}. There exists a trade-off between the quantity of data used and the accuracy of the corresponding pseudo labels~\cite{subset_ws_neurips22}. Generally, it is customary to utilize the entire dataset for training purposes. However, this approach may introduce unnecessary noise and potentially hinder the overall performance of the classifier. A possible solution is to formulate the given problem as a subset selection problem and select the most representative samples with the least entropy associated with the pseudo labels for the training of the classifier~\cite{subset_ws_neurips22}. However, these methods do not provide any guarantee for the representativeness and diversity of the samples used for training and are not computationally suited for an adaptive training setting, which requires selecting multiple pseudo-labeled subsets. 
\par Selecting an optimal subset of representative samples from the training data is an NP-hard problem. However, if the function used to measure the utility of a given set is shown to be submodular, then the set generated by maximizing the given utility using a greedy based selection scheme with constraints like cardinality~\cite{nemhauser1978analysis} and knapsack~\cite{submodular_knapsack_NAS,leskovec2007cost,krause_submodular_maximization}, is guaranteed to be very close to the optimal set, thereby approximately solves the subset selection problem. 
% \begin{definition}
%     A function $\gF: 2^{\gV} \rightarrow R$, where $\gV$ is a finite set is submodular, if for every $\gA \subseteq \gB \subseteq \gV$ and $\forall$ $\vk \in \gV \setminus \gB$ $\gF(\vk \bigcup \gA) - \gF(\gA) \geq \gF(\vk \bigcup \gB) - \gF(\gB)$
% \end{definition}
% Intuitively, this is similar to the diminishing return property, where with growing size the overall gain of adding any new element in the set decreases. Further, a specific class of submodular function is called monotone if $\forall (\gA, \gB \subset \gV) $  s.t.  $\gA \subseteq \gB ,$   $ \gF(\gA) \leq \gF(\gB)$

% Generally, a submodular optimization of the form $\max_{S\subset \gV} \gF(S) $ under basic constraints defined on S like the cardinality constraint. i.e., $|S| \leq L$  is NP-hard. However, it has been shown that a normal greedy algorithm can generate a good subset which is very close to optimal. Under a greedy setting for a normal cardinality constraint, a subset can be generated by starting with an empty set ($\gS_0$) and iteratively adding data point $ v = argmax _{ v \in \gV \setminus \gS_{n-1}  } \gF(v \bigcup \gS_{n-1}) - \gF(\gS_{n-1})$ where ($\gS_n = \gS_{n-1} \bigcup {v}$). Studies have shown that  a subset selected by such a greedy selection procedure is very close to the optimal~\citep{nemhauser1978analysis}.  $\gF(\gS_{r}) \geq (1-\frac{1}{e}) \max_{|\gS| \leq R} \gF(S)$. knapsack constraint~\cite{krause_submodular_maximization} is a special case of submodular optimization where there are different cost associated with adding every new element in the set.

Recent works~\cite{submodularity_AI_blimes,submodular_knapsack_NAS,joseph2019submodular,simsar2023fantastic,wei2015submodularity,maheshwari2021learning,sinha2020small}, have used submodular functions to select a subset of the entire data for efficient training of deep learning models.  

%\begin{figure*}
%\centering
%\resizebox{0.90\textwidth}{!}{ 
  %  \includegraphics{AAAI/AnonymousSubmission-2024/Architecture_pws.png}
    
  %  \caption{Given diagram depicts the complete training process of our proposed method. The overall training consist of three major components. (1) A classifier is trained using the subset of data selected by our submodular maximization scheme,(2) secon}
%\end{figure*}
\section{Proposed Method} 
\subsection{Problem setting and overview}
Let there be $n$ training data samples $ \Tilde{\gD} = \{x_1, x_2, \hdots x_n\}$ drawn i.i.d from a distribution $\gP_{x}$. The objective of our formulation is two folds: firstly, we want to infer the class labels for these samples i.e., $y \in \{0 \hdots \gC\}$ and secondly, we want to sample synthetic images whose marginal is represented as $\gP_{x}$. In our setting, instead of observing true labels, we have access to the predictions of  $\gK$ label functions on given training data. Each label function $\lambda_{k} (x_i)$ (where $k \in \{1 \hdots\gK\}$) generates a noisy prediction of the true class label for a sample $x_{i}$. These label functions can either predict among a given set of possible class labels i.e, $\lambda_{k} (x_i) \in \{0 \hdots  \gC\}$ or can abstain from making a prediction because of low confidence~\cite{PWS_ratner2016}. Let us refer to the data with a prediction from at least one label function as $\Tilde{\gD_t} \in \Tilde{\gD}$ (non-abstained dataset). For each data sample we have an associated prediction from $\gK$ label functions represented as $\Lambda(x_i) = (\lambda_1(x_i), \lambda_2(x_i) \hdots \lambda_{\gK}(x_i))$. A label model 
 ($\gL$) utilizes the predictions made by these  LFs to predict a final pseudo label ($\hat{y_i}$) for a given input sample $x_i$. To train our noise-aware classifier, we generate a subset of original training data $\Tilde{\gD_0} \in \Tilde{\gD_t}$ and their associated pseudo labels ($\hat{\gY} = \{\hat{y_1} \hdots \hat{y}_{|\Tilde{\gD}_0|}$ \}) and study the impact of training the noise-aware classifier with the given subset on the overall performance of the conditional GAN and the label model. 

 \begin{figure}[H]
    \centering
\includegraphics[width=0.23\textwidth]{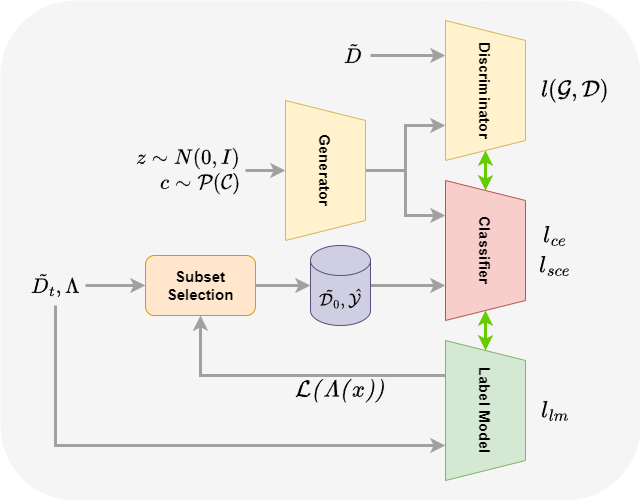}
    \caption{Overview of the proposed fusion between conditional GAN and Programmatic Weak Supervision. In the current setup, a subset of the non-abstained samples, selected by submodular maximization under a knapsack constraint, is used to train a noise-aware classifier. This classifier shares its network weights with the discriminator and label model. The noise-aware classifier is then employed to train a conditional GAN and refine the label model. As the training progresses and the label model becomes more accurate, a better subset is utilized to train the classifier. }
    \label{fig:arch}
\end{figure}

 \subsection{Noise-aware classifier training}
As stated previously, we have used a classifier to facilitate the training of the conditional GAN and the alignment of the label model. Given the absence of ground truth labels, the classifier is trained using the pseudo labels generated by the label model. 

The cross-entropy loss is a common choice of the loss function used to train a classifier. However, recent studies~\cite{symmetric_cross_entropy,feng2021can} have shown that a classifier trained only with cross-entropy loss under a noisy setting tends to exhibit prediction bias and is susceptible to overfitting. In the context of weak supervision, such a classifier can not only impact the performance of the conditional GAN but can also lead to further deterioration of the quality of pseudo labels generated by the label model.

In our study, we address this issue by training our classifier using a symmetric cross-entropy loss ($\vl_{sce}$)~\cite{symmetric_cross_entropy} between the pseudo label ($\hat{y} \sim \hat{\gY}$) generated by the label model, and the class conditional distribution modeled by the classifier ($\gP(t|x)$). The classifier is trained on a subset of the training dataset ($\Tilde{\gD}_0$) selected by our proposed submodular maximization scheme. Symmetric cross-entropy loss is a weighted combination of standard cross-entropy loss ($\vl_{ce}$) and reverse cross-entropy loss ($\vl_{rce}$). While the cross-entropy loss provides a better convergence to the classifier, reverse cross-entropy loss provides robustness to different types of noise such as uniform and class conditional noise~\cite{symmetric_cross_entropy,robust_classifier_ghosh}, thereby ensuring that the classifier does not overfit to noisy class labels. Mathematically, the symmetric cross-entropy loss ($\vl_{sce}$) is described in Eq. \ref{eqn:classifier}.

\begin{align}
\vl_{ce}(x, \hat{y})  & =   -\sum_{t=0}^{\gC} \hat{y}_t \log(\gP(t|x)) \nonumber\\
\vl_{rce} (x, \hat{y})  & = -\sum_{t=0}^{\gC} \gP(t|x)\log(\hat{y}_t) \nonumber\\
\vl_{sce}(x, \hat{y}) & = \alpha \vl_{ce}(x, \hat{y}) + \beta \vl_{rce} (x, \hat{y}) \nonumber 
\\
\vl_{sce}^{\Tilde{\gD_0}} & = \frac{1}{|\Tilde{\gD_0}|} \sum_{x\sim \Tilde{\gD_0}, \hat{y} \sim \hat{\gY}} \vl_{sce}(x, \hat{y})
 % \end{autobreak}
 \label{eqn:classifier}
\end{align}

The classifier is trained in an adaptive manner ~\cite{psuedolabel_gan_Morerio}, where a new set of data ($\Tilde{D}_0$) is introduced for training after a specific number of epochs. During the training, the given examples and their associated pseudo labels remain unchanged until the next iteration. This ensures that as the accuracy and performance of the label model improve, a more refined dataset and its corresponding pseudo labels are utilized for training. Simultaneously, it prevents the model from diverging due to frequent alterations in data samples and their associated pseudo labels. 
\subsection{Conditional GAN}
In our study, we have employed a classifier-guided generative adversarial network (GAN) to generate class conditional images~\cite{ACGAN}. The generator ($\gG$) is fed a noise sample $z\sim N(0,\textit{I})$ and a one hot representation of class label $c\sim \gP(\gC)$  to sample an image ($x_g$ = $\gG(z,c)$) from the parameterized distribution ($\gP_\theta$). The discriminator ($\mathcal{D}$) tries to distinguish between the images sampled from real data distribution ($\gP_x$) and the images sampled from parameterized distribution ($\gP_{\theta}$). For conditional image generation, the generator further maximizes the likelihood of the image ($x_g$) belonging to class c with respect to our noise-aware classifier. The objective function that is optimized for Conditional GAN training is as follows:

\begin{align}
 \vl(\gG, \gD) =& \underset{x\sim \gP_x}{\E}\left[\log (\gD(x))\right] + \underset{x\sim \gP_\theta}{\E}\left[\log(1-\gD(x))\right] \nonumber \\
 \min_{\gG} \max_{\gD} \vl_{gan} =& \vl(\gG, \gD) + \underset{x\sim \gG(z,c), c \sim \gP(\gC)}{\E}\left[\vl_{ce} (x,c)\right]
    % \min{\gG}\max_{\gD} \v_{gan} =  \gL_{gan} - \lambda \left[\underset{\hat{\gY} \sim \gL(\Lambda(x)), x\sim \gP_x}{\E}\left[ \gL_{\textit{sce}}(\hat{\gY}, \gC(x))\right]\right] \\
    %  \min_{\gG} V(\gG,\gD) + \lambda\left[\underset{c \sim \gP(c), x\sim \gP_\theta}{\E}\left[\gL_{\textit{ce}} (c, \gC(x))\right]\right]
    % \\
    % \underset{c \sim \gP(c), x\sim \gP_\theta}{\E}\left[\gL_{\textit{ce}}\right]  =  \frac{1}{\gB}\sum_{x}\sum_{t=0}^{\gM} c_{t}log(\gC(t|x)) 
\end{align}

\subsection{Construction of label models}
\label{sec:labelmodel}
In programmatic weak supervision, the label model accumulates the predictions made by different label functions to make an informed final estimation of the pseudo label associated with the given training sample. Generally, this is done by defining an accuracy potential function ($\phi_k$) to capture the association between the prediction made by a given label function and any class label $y$ ($\phi_k(\lambda_k,y)$). Similar to previous works \cite{gen_model_PWS_vice_versa, PWS_ratner2020snorkel}, we have defined the accuracy potential as an indicator function, $\phi_k(\lambda_k,y) = \vone(\lambda_k = y)$ which captures the notion whether a given label function ($\lambda_k$) has predicted class $y$ for a given input sample. With each label function, there is an associated accuracy parameter ($\gA_k: \gR^N\rightarrow (0,1)$), which determines the accuracy of its prediction. We have used sample-dependent accuracy parameters, where the accuracy of every such association is different for different input samples~\cite {cauchy_end_to_end, gen_model_PWS_vice_versa}. The label model ($\gL({\Lambda(x)})_j$), defined below, captures the probability associated with class $j$  for any input sample $x$.
\begin{align}
    \gL({\Lambda(x)})_j = \frac{\exp(\sum_{i=0}^{\gK}\gA(x)_i \phi_i(\lambda_i(x), j))}{\sum_{\vs \in y} \exp(\sum_{i=0}^{\gK} \gA(x)_i \phi_i(\lambda_i(x), \vs)) } \nonumber \\
    \text{ where } j\in \{ 0 \hdots \gC\} 
\end{align}

The final pseudo label, associated with the input sample $x$, is defined as the class with the maximum probability.   
 
\begin{align}
\hat{y} = \argmax_j \gL({\Lambda(x)})_j  
\end{align}
% \end{equation}

To enhance the precision of the generated pseudo labels, the label model is trained using the probabilities generated by the classifier for the samples from the non-abstained dataset ($\Tilde{\gD}_t$). In this approach, we employ a linear layer followed by a softmax ($\hat{\gF}$) for the alignment of probabilities between the output of the label model and the classifier ($\gP(t|x)$)~\cite{gen_model_PWS_vice_versa}, giving rise to the following loss function. 
\begin{align}
    \centering
    \vl_{align}(x) =  -\sum_{t=0}^{\gC} \gP(t|x)\log(\hat{\gF}( \gL(\Lambda(x))_t)
\end{align}
Since our classifier is trained iteratively by utilizing the pseudo labels generated through the label model  $\big(\hat{y}$ in Eq.~\ref{eqn:classifier}$\big)$, we ensure that the pseudo labels are not entirely random, especially during initial epochs. To accomplish this, we employ a new loss $\vl_{decay}$  that assigns equal weights to all the accuracy parameters and gradually decay this loss as the label model becomes more accurate~\cite{gen_model_PWS_vice_versa}. Under this loss, the accuracy parameter for all the label functions is approximated to a value of $0.5$, and the overall loss is decayed by a rate $\mu$ as the training proceeds. Mathematically, 
it is equivalent to generating an accuracy parameter of 0.5 for all the label functions, which is approximated by a vector $0.5 *\overrightarrow{\vone}$ where $\overrightarrow{\vone}$ is a vector of ones  with dimension equal to the number of label functions, giving rise to the following loss function :
\begin{align}
    \centering
    \vl_{decay}(x) = & \mu||\gA(x)- 0.5 *\overrightarrow{\vone}||^2
\end{align}

This process ensures that during the initial epochs, the label model approximates a majority vote, and as training progresses, it becomes more data-specific.  
Finally, the label model is trained using 
 $\vl_{lm}$, by combining  alignment and decay loss $\vl_{lm} =  \vl_{align} + \vl_{decay}$.

\subsection{Subset 
selection for classifier training}\label{sec:subset_selection}
To facilitate better training of our noise-aware classifier, we select a subset of data ($\Tilde{D}_0$) from the training examples that have received at least one vote from the label functions ($\Tilde{D}_t$). We formulate this as a submodular optimization problem as in ~\cite{submodularity_AI_blimes,mirzasoleiman2013distributed}. 

\subsubsection{Preliminary on submodularity}
\begin{definition}
    A function $\gF: 2^{\gV} \rightarrow R$ is said to be submodular for a finite set $\gV$,  if for every $ A \subseteq B \subseteq \gV$ and $\forall$ $v \in \gV \setminus B$ $\gF\big(A \bigcup v\big) - \gF\big(A\big) \geq \gF\big(B \bigcup v\big) - \gF\big(B\big)$
    \label{Defn:submod}
\end{definition}
%The definition of submodular function (Def-\ref{Defn:submod}) is similar to the diminishing return property, where with growing size, the overall gain of adding any new element in the set decreases. 
Further, a specific class of submodular functions is called monotone if $\forall \big(A, B \subset \gV\big) $  s.t.  $A \subseteq B,$   $ \gF\big(A\big) \leq \gF\big(B\big)$.

% The problem of finding the optimal solution for a submodular maximization objective is known to be NP-hard. Therefore, none of the existing algorithms can guarantee the discovery of an optimal solution for given objective. In such scenarios, researchers often rely on designing heuristics like greedy algorithm to generate the subset.

A  monotonous submodular function can be used to determine the utility of a given set of data for a downstream task, and an efficient subset of the data can be selected by maximizing the given function. For a cardinality constraint, the process begins with an empty set ($\gS_0$), and the selection proceeds iteratively in a greedy manner, selecting the data point that maximizes the submodular function until the given budget on the size is exhausted. Mathematically, if $v$ denotes the datapoint to be selected, then $ v = \argmax _{ v \in \gV \setminus \gS_{n-1}  } \gF\big(v \bigcup \gS_{n-1}\big) - \gF\big(\gS_{n-1}\big)$  where $\big(\gS_n = \gS_{n-1} \bigcup {v}\big)$. Although this heuristic method doesn't guarantee the generation of an optimal subset due to the NP-hard nature of the problem, it can still yield a solution that is assured to be in proximity to the optimal set~\cite{nemhauser1978analysis} and hence approximately solves the submodular maximization problem. Knapsack constraint~\cite{krause_submodular_maximization} is another special case of submodular maximization, where there are different costs associated with adding every new element.

% Generally, a submodular optimization of the form $\max_{S\subset \gV} \gF(S) $ under basic constraints defined on S like the cardinality constraint. i.e., $|S| \leq L$  is NP-hard. However, it has been shown that a normal greedy algorithm can generate a good subset which is very close to optimal. Under a greedy setting for a normal cardinality constraint, a subset can be generated by starting with an empty set ($\gS_0$) and iteratively adding data point $ v = argmax _{ v \in \gV \setminus \gS_{n-1}  } \gF(v \bigcup \gS_{n-1}) - \gF(\gS_{n-1})$ where ($\gS_n = \gS_{n-1} \bigcup {v}$). Studies have shown that  a subset selected by such a greedy selection procedure is very close to the optimal~\citep{nemhauser1978analysis}.  $\gF(\gS_{r}) \geq (1-\frac{1}{e}) \max_{|\gS| \leq R} \gF(S)$. 

\subsubsection{A submodular framework for subset selection}

% \par To facilitate the training of our noise aware classifier, we select a subset of data ($\Tilde{D}_0$) from the training examples that have received at least one vote from the label functions ($\Tilde{D}_t$). 
% This is essential because, without at least a single vote, the label model won't be able to generate a pseudo label for those specific examples.
% The training examples in the selected dataset is accompanied with the pseudo labels($\gY$) generated by the label model. 
 In our work, we have used a generalized graphcut-based monotonous submodular function~\cite{iyer_graph_cut,submodularity_AI_blimes}  that uses a greedy approach to select a set of representative and diverse examples to train the classifier. Further, the selection is done under a knapsack constraint where the addition of any new data sample has a cost equal to the entropy of the pseudo label associated with it, and the selection of samples is done under a budget on the overall entropy of the subset. Eq.~\ref{eqn:sub} illustrates the formulation of the generalized graphcut algorithm. The function $(\gF\big(S\big))$ represents the utility of the set $(S)$, and the variable $s_{i,j}$ represents a kernel function that quantifies the similarity between the $i^{th} \text { and } j^{th}$ samples in the dataset. We have used cosine similarity over the Inception features~\cite{devries2020instance} of the images to define our kernel function. The term $\big(\sum_{i\in \Tilde{\gD}_t} \sum_{j \in \gS} \vs_{ij}\big)$ represents a cumulative score for the similarity between the samples in set $(S)$ and the data present in the non-abstained dataset ($\Tilde{\gD}_t$), capturing the representativeness of the selected samples in comparison to $\big(\Tilde{\gD}_t\big)$. Additionally, the term $\big(\sum_{l,m \in \gS}  \vs_{l,m}\big)$ represents the similarity between the selected samples in set $\big(S\big)$, minimizing which, leads to the selection of diverse data points. The hyperparameter $\gamma$ controls the tradeoff between representativeness and diversity. Further, choosing $\gamma \geq 2$ ensures that the given function exhibits monotonicity.    
\begin{align}
\gF\big(\gS\big)   = \gamma \sum_{i\in \Tilde{\gD}_t} \sum_{j \in \gS} \vs_{ij} - \sum_{l,m \in \gS} \vs_{l,m}   
\label{eqn:sub}
\end{align}
We have defined a knapsack constraint on the overall entropy of the pseudo labels generated by label model for the selected samples. Under the given constraint, every data sample in $\Tilde{\gD_t}$ has an associated selection cost equal to the entropy of the pseudo label, and the overall entropy of the selected samples cannot exceed a given budget ($\gB$) (Eq.~\ref{eqn:budget}). This enforces the selection of diverse and representative samples while ensuring that uncertainty associated with the pseudo labels for the selected samples is within a limit. The complete objective of our formulation is stated as follows: 
\begin{align}
    \max_{\gS \subset \Tilde{\gD}_t} \gF\big(\gS\big) \nonumber \\
    \text {s.t. Cost ($\gS$)} \leq \gB 
    \label{eqn:budget}
\end{align}
The budget $\gB$ on the overall entropy of the pseudo labels for the selected samples is defined to be a fraction ($\eta \text{ where } 0 < \eta < 1$) of the total entropy of the pseudo labels ($Cost(\Tilde{\gD_t})$) for the non-abstained data set. Eq.~\ref{Eqn:cost} illustrates the overall constraint on the given submodular maximization problem, where $\gH$ is the entropy defined for a given sample ($x_l$), and $\gL(\Lambda(x_l))$ is the probability distribution of the label model associated with it.
\begin{align}
    Cost(\gS) &= \sum_{x_l \in \gS} \gH\big(x_l; \gL(\Lambda(x_l))\big) \nonumber \\ 
        Cost(\Tilde{\gD_t}) &= \sum_{x_l \in \Tilde{\gD_t}} \gH\big( x_l; \gL(\Lambda(x_l))\big) \nonumber \\
        \gB &= \eta*Cost(\Tilde{\gD_t}) 
        \label{Eqn:cost}
\end{align}

For the submodular maximization problem under a knapsack constraint, a naive greedy algorithm can perform arbitrarily bad as it is indifferent to the cost associated with each sample. To account for the selection cost associated with every sample, the normal greedy-based selection scheme can further be modified to create a cost-effective greedy algorithm (CEG) (Eq.~\ref{eqn:ceg})
to select the optimal set, where $c(.)$ is the cost function defined for every input sample, $v_k$ is the data point that is selected at $k^{th}$ iteration, $\gF\big(\gS_{k-1}\big)$ is the utility of the set ($S_{k-1}$) as per the the submodular function defined in Eq.~\ref{eqn:sub} and $\gS_k$ is the new set created by including the data point $\vv_k$ in the set $\gS_{k-1}$. 
In our formulation, the function $c(.)$ is set to be equal to the entropy associated with the pseudo label generated by label model $\big(c(v) = \gH(v;\gL(\Lambda(v)))\big)$. The cost-effective greedy can be summarized as below :        
\begin{align}
    v_k = \argmax_{v\in \Tilde{\gD_t}/\ \gS_{k-1}} \frac {\gF\big(\gS_{k-1} \cup {v}\big) - \gF\big(\gS_{k-1}\big)} {c(v)} \nonumber \\
    \text{ where } \gS_{k} = \gS_{k-1} \bigcup v_{k}
    \label{eqn:ceg}
\end{align}
The given algorithm is run iteratively until the total cost of the selected set is less than the predefined budget ($\gB$).

Even though cost-effective greedy considers the selection cost for each sample while selecting a subset, it can still perform arbitrarily bad in some extreme scenarios~\cite{leskovec2007cost}. Fortunately,
 for the given submodular maximization objective, the cost-effective greedy algorithm can be adapted to provide a constant factor of $\frac{1}{2} \big(1- \frac{1} {e}\big)$ to the optimal solution (Prop. 1 in supplementary material)~\cite{leskovec2007cost, submodular_knapsack_NAS}, hence generating a subset close to the optimal set. In fact, at least one among the 
 final set ($\gS_{uniform}$) selected by using uniform selection cost ($(c(v) = 1)$) and the final set ($\gS_{cost}$) selected using desired cost function $\big(c(v) = \gH(v;\gL(\Lambda(v)))\big)$
 % one of the methods with uniform cost $\gF(\Tilde{\gS}_{Uniform})$ ($(c(v) = 1)$) and one generated with desired cost function $\gF(\Tilde{\gS}_{Cost})$ $(c(v) = \gH(v_l;\gL(\Lambda(v))))$
 will have a constant approximation bound with respect to the optimal solution, and the set with the maximum utility score can be used for the training of the classifier \big($\Tilde{\gD}_0 = \argmax_{\gS} (\gF(\gS_{uniform}),\gF(\gS_{Cost}) \big)$.
 % \begin{prop} \cite{leskovec2007cost,submodular_knapsack_NAS}
 %     If $\mathcal{\gF}$ is a non decreasing submodular function with $\gF\big(\phi\big)$ = 0 then CEG algorithm achives a constant ratio of  $\frac{1}{2} \big(1- \frac{1}{e}\big)$  to the optimal solution under the knapsack constraints.   
 %     \begin{align}
 %     \max \big(\gF(\Tilde{\gS}_{Uniform}),
 %     \gF(\Tilde{\gS}_{Cost})\big)\geq \frac{1}{2} \big(1- \frac{1}{e}\big)\max_{Cost(A)\leq \gB } \gF(\gS) \nonumber
 %     \end{align}
 %     \label{prop:submod}
 % \end{prop}
 
 % where $\Tilde{\gS}_{Cost}$ and $\Tilde{\gS}_{Uniform}$ are optimal subsets generated by using cost effective greedy algorithm   (Eq.~\ref{eqn:ceg}) for entropy based cost and uniform cost. The subset with the maximum value for the utility function is used for the training of the classifier ($\Tilde{\gD}_0 = \argmax_{\Tilde{\gS}} (\gF(\Tilde{\gS}_{uniform},\Tilde{\gS}_{Cost} ))$).

\section{Experiments}
We assess the effectiveness of our method by experimenting with it on different datasets and label functions. Specifically, we use the label functions provided by the authors of WSGAN~\cite{gen_model_PWS_vice_versa}. The primary experiments are done on five datasets, namely Animals with Attributes 2 
 (AWA2)~\cite{Awa2}, DomainNet~\cite{domainnet}, CIFAR10~\cite{CIFAR10}, MNIST~\cite{MNIST}, FashionMNIST~\cite{FMNIST} and German Traffic Sign Benchmark (GTSRB)~\cite{GTSRB}. These experiments are conducted on the following four types of label functions:

\begin{table*}[htp!]
    \centering
     \resizebox{0.7\textwidth}{!}
 {
        \begin{tabular}{c|ccccccccc}
            \hline
            \textbf{Dataset} & \textbf{MV} & \textbf{MeTaL} & \textbf{FS} & \textbf{Snorkel} & \textbf{DS} & \textbf{HLM}& \textbf{WSGAN} & \textbf{Ours (comp)} & \textbf{Ours (sub)} \\ \hline
            AWA2 & 0.627 & 0.622 & 0.621 & 0.617 & - & 0.651 &0.670 & \textbf{0.701} & 0.698 \\ 
            CIFAR10-A & 0.831 & 0.804 & 0.800 & 0.804 & 0.850& 0.842 & 0.872 & \textbf{0.890} & \textbf{0.890} \\ 
            CIFAR10-B & 0.719 & 0.708 & 0.708 & 0.709 & 0.677 & 0.726 & 0.729 & \textbf{0.740} & \textbf{0.740} \\ 
            DomainNet & 0.595 & 0.486 & 0.635 & 0.507 & \textbf{0.658} &0.646 & 0.643 & 0.650 & 0.653 \\ 
            FashionMNIST & 0.735 & 0.730 & 0.734 & 0.725 & 0.717 & 0.735 & 0.744 & \textbf{0.754} & \textbf{0.754} \\ 
            GTSRB & 0.816 & 0.619 & 0.815 & 0.671 & 0.814 & - & 0.825 & \textbf{0.829} & 0.828 \\ 
            MNIST & 0.778 & 0.758 & 0.773 & 0.759 & 0.729 & 0.781 & 0.816 &  0.818 & \textbf{0.819} \\ \hline
        \end{tabular}
    }
    \caption{Comparison between the average posterior accuracy of the label models for samples with at least one vote from the label function. The best performer across methods is denoted as \textbf{bold}.}
    \label{tab:label_model_acc}
\end{table*}
\begin{table}[ht!]
    \centering
    \resizebox{0.7\linewidth}{!}
{
    \begin{tabular}{c|ccc}
    \hline
        \textbf{Dataset} & \textbf{WSGAN} & \textbf{Ours (comp)} & \textbf{Ours (sub)} \\ \hline
        AWA2 & 36.00  & 25.92 & \textbf{25.23} \\
        CIFAR10-A & 22.71 & 18.00 & \textbf{17.50} \\ 
        CIFAR10-B & 24.41 & \textbf{18.72} & \textbf{18.72} \\ 
        DomainNet & 44.35 & 37.42 & \textbf{36.82} \\ 
        GTSRB & 73.96  & 62.55 & \textbf{59.09} \\ 
        FashionMNIST & 15.94 & \textbf{11.78} & 11.81 \\
        MNIST & 5.35 & \textbf{4.00} & \textbf{4.00} \\
 \hline
    \end{tabular}
}
    \caption{Comparison of image quality (mean FID) score of the proposed method with WSGAN. The best performer across all methods is denoted as \textbf{bold}.}
    \label{Tab:FID}
\end{table}

\begin{table}[ht!]
\centering
\resizebox{0.50\textwidth}{!}
{
\begin{tabular}{c|cc|cc|cc}
\hline
\textbf{Dataset}& \textbf{WG-S} & \textbf{WG-LF} & \textbf{Ours(S)}-L & \textbf{Ours(s)-L} & \textbf{Ours(C)-CL} & \textbf{Ours(S)-CL} \\
\hline
AWA2& 0.99 & 0.99 & \textbf{2.58} & 1.59 & 1.59 & \textbf{1.79} \\
CIFAR10-A& 1.38 & 0.46 & 0.50 & \textbf{0.76} & 1.74 & \textbf{2.16} \\
CIFAR10-B& 0.66 & 0.36 & 0.42 & \textbf{0.82} & 1.22 & \textbf{1.56} \\
DomainNet& 0.84 & 0.84 & 1.88 & \textbf{3.14} & \textbf{3.77} & 1.88 \\
FashionMNIST& 0.36 & 2.72 & 2.74 & \textbf{2.92} & \textbf{0.74} & 0.64 \\
GTSRB& 0.05 & 0.32 & \textbf{1.06} & 0.60 & 0.81 & \textbf{1.27} \\
MNIST& 0.30 & 0.08 & \textbf{0.34} & 0.22 & \textbf{0.52} & 0.38 \\
\hline

\end{tabular}
}
\caption{Percentage increase in the test accuracy of ResNet-18 using synthetic data augmentation. As a baseline, the WSGAN model with synthetic pseudo labels (WG-S) and label function-based pseudo labels (WG-LF) is used. For the proposed methods, we have compared the complete data setting (Ours(C)) and subset selection setting (Ours(S)) using conditional latent vectors (L) and classifier-based (CL) pseudo labels. The best performer for a given pseudo label is highlighted in \textbf{bold.}}
\label{tab:resnet}
\end{table}

 \textbf{Domain transfer}: Experiments related to DomainNet use a domain transfer framework, wherein the label functions are trained using images from distinct source domains, like paintings, art and are subsequently used to generate weak labels for a new target domain like real-world images~\cite{mazzetto2021_domainnet}.
\textbf{Attribute heuristics}: Experiments related to AWA2 use attribute classifiers that are trained on a fixed set of animal images. The weak labels for the unseen dataset are generated by training a decision tree on top of the predictions made by these classifiers. 
\textbf{Self-supervised learning}: Experiments related to CIFAR10-B, MNIST, GTSRB, and FashionMNIST use label functions generated by finetuning a shallow MLP network over a small validation dataset. A SimCLR ~\cite{SIMCLR}  model pre-trained on ImageNet was utilized to generate features for this experiment.  
    % This basically entails fine tuning experiments where the features is learnt using SimCLR~\cite{SIMCLR} on ImageNet. A MLP based classifer is trained using a small validation data which is then subsequently used to generate weak labels for held out dataset. Experiments related to CIFAR10-B, MNIST and FashionMNIST are done using label function generated by given approach. 
\textbf{Synthetic}: Experiments related to CIFAR10-A  are conducted using label functions generated by synthetically corrupting the true class labels of data under given a propensity and accuracy rate.
Further details about the label functions and their associated accuracy and coverage can be found in WSGAN~\cite{gen_model_PWS_vice_versa}. 

\subsection{Baselines and evaluation metrics}
We analyze the performance of two versions of our model. Firstly, we conduct experiments by considering all the samples where the label function has provided at least one vote for the classifier's training i.e., $\Tilde{\gD_0} = \Tilde{\gD_t}$. The associated results related to these experiments are reported as \textit{Ours (comp)}. Secondly, we conduct experiments where the dataset for classifier training is selected using the proposed submodular scheme. The results for these experiments are reported as  \textit{Ours (sub)}. We used Apricot library~\cite{apricot} to perform submodular optimization for subset selection using the lazy greedy method. All the major experiments are conducted using a DCGAN~\cite{DCGAN} architecture. In the current design, the discriminator network shares weights with the classifier and accuracy parameter-based model. Further, we have provided a network ablation over styleGAN-ADA~\cite{styleGAN-ADA} architecture under a similar configuration for CIFAR10-B and high-resolution images in the supplementary material, which also includes the hyperparameters of subset selection and other implementation details. 

We utilized the FID~\cite{FID} score to assess and compare the image quality produced by our approach. Similar to WSGAN, we conducted five independent iterations of our proposed technique and recorded the average FID score. Additionally, as PWS operates in a transductive setting, we follow the prior work and have reported the mean accuracy of the pseudo labels on the training data to compare the performance of the label model with other baselines. Metrics related to the F1 score, mean precision, and other related scores are provided in the supplementary material. We have also evaluated the performance of the proposed method on the downstream data augmentation task.

\begin{figure}[ht!]
    \centering
    \resizebox{0.42\textwidth}{!}
 {
    % This file was created with tikzplotlib v0.10.1.
\begin{tikzpicture}

\definecolor{darkgray176}{RGB}{176,176,176}
\definecolor{darkorange25512714}{RGB}{255,127,14}
\definecolor{forestgreen4416044}{RGB}{44,160,44}
\definecolor{lightgray204}{RGB}{204,204,204}
\definecolor{steelblue31119180}{RGB}{31,119,180}

\begin{axis}[
legend cell align={left},
legend style={
  fill opacity=0.8,
  draw opacity=1,
  text opacity=1,
  at={(0.97,0.03)},
  anchor=south east,
  draw=lightgray204
},
tick align=outside,
tick pos=left,
title=\Huge\textbf{GTSRB},
x grid style={darkgray176},
xlabel=\Huge{Epoch},
xmajorgrids,
xmin=-9.95, xmax=208.95,
xtick style={color=black},
y grid style={darkgray176},
ylabel=\Huge{ARI},
ymajorgrids,
ymin=-0.03415, ymax=0.8,
ytick style={color=black},
font=\Huge,
xlabel style={yshift=-14pt},
ylabel style={yshift=14pt}
]
\addplot [line width=3, steelblue31119180]
table {%
0 0.0022
1 0.2449
2 0.5064
3 0.5718
4 0.614
5 0.637
6 0.6524
7 0.6649
8 0.6687
9 0.6776
10 0.6779
11 0.6849
12 0.6867
13 0.6913
14 0.69
15 0.7076
16 0.6988
17 0.6978
18 0.705
19 0.6991
20 0.7085
21 0.7116
22 0.7113
23 0.7094
24 0.718
25 0.7182
26 0.7138
27 0.7153
28 0.7193
29 0.7131
30 0.7177
31 0.7247
32 0.7237
33 0.7151
34 0.7204
35 0.7177
36 0.7239
37 0.721
38 0.7272
39 0.7192
40 0.7205
41 0.7246
42 0.7152
43 0.715
44 0.7227
45 0.7197
46 0.7185
47 0.7207
48 0.7234
49 0.72
50 0.7229
51 0.7183
52 0.7208
53 0.7218
54 0.7178
55 0.7292
56 0.7185
57 0.7226
58 0.7174
59 0.7261
60 0.7286
61 0.7256
62 0.7238
63 0.7255
64 0.7207
65 0.7237
66 0.7193
67 0.7199
68 0.7194
69 0.7175
70 0.7188
71 0.7212
72 0.7185
73 0.7238
74 0.7246
75 0.7204
76 0.7195
77 0.7247
78 0.7249
79 0.7217
80 0.7183
81 0.7211
82 0.7219
83 0.726
84 0.7281
85 0.7208
86 0.717
87 0.723
88 0.7259
89 0.72
90 0.7219
91 0.7248
92 0.7249
93 0.7228
94 0.7226
95 0.7258
96 0.7234
97 0.7259
98 0.7258
99 0.7209
100 0.7258
101 0.7222
102 0.7175
103 0.7242
104 0.7242
105 0.7243
106 0.719
107 0.727
108 0.722
109 0.7209
110 0.7226
111 0.7242
112 0.7238
113 0.7238
114 0.7227
115 0.7234
116 0.7209
117 0.7285
118 0.7276
119 0.7221
120 0.7215
121 0.7224
122 0.7229
123 0.7207
124 0.7254
125 0.7234
126 0.7272
127 0.7243
128 0.7191
129 0.7205
130 0.7254
131 0.7245
132 0.7199
133 0.723
134 0.7195
135 0.7227
136 0.7239
137 0.7175
138 0.719
139 0.7205
140 0.723
141 0.7263
142 0.7229
143 0.7242
144 0.7214
145 0.7229
146 0.724
147 0.7236
148 0.7223
149 0.7178
150 0.722
151 0.7233
152 0.7203
153 0.7242
154 0.7222
155 0.7255
156 0.7269
157 0.7215
158 0.7198
159 0.7244
160 0.7226
161 0.7238
162 0.7249
163 0.7223
164 0.7231
165 0.7241
166 0.7189
167 0.7219
168 0.7233
169 0.7182
170 0.7232
171 0.7212
172 0.7228
173 0.7239
174 0.7254
175 0.7246
176 0.7232
177 0.7226
178 0.7204
179 0.7262
180 0.7271
181 0.7262
182 0.722
183 0.7225
184 0.7236
185 0.7175
186 0.7249
187 0.7249
188 0.7199
189 0.7227
190 0.7221
191 0.7208
192 0.7217
193 0.7203
194 0.7221
195 0.7226
196 0.7217
197 0.7244
198 0.7184
199 0.7212
};
% \addlegendentry{Ours(complete data)}
\addplot [line width=3, darkorange25512714]
table {%
0 0.0071
1 0.2671
2 0.516
3 0.5808
4 0.6225
5 0.6359
6 0.6567
7 0.6626
8 0.6689
9 0.6767
10 0.6843
11 0.6861
12 0.6807
13 0.689
14 0.693
15 0.6843
16 0.6916
17 0.6908
18 0.6965
19 0.6964
20 0.7027
21 0.7017
22 0.7028
23 0.7068
24 0.7034
25 0.7007
26 0.7011
27 0.7047
28 0.708
29 0.7045
30 0.7087
31 0.7085
32 0.7014
33 0.7074
34 0.7116
35 0.7079
36 0.7066
37 0.71
38 0.7093
39 0.715
40 0.7084
41 0.7053
42 0.7108
43 0.7051
44 0.7091
45 0.7042
46 0.7132
47 0.7105
48 0.7069
49 0.7109
50 0.71
51 0.706
52 0.7084
53 0.707
54 0.7124
55 0.7109
56 0.7107
57 0.7173
58 0.7098
59 0.7135
60 0.7183
61 0.7188
62 0.7117
63 0.7167
64 0.7086
65 0.7059
66 0.7125
67 0.7122
68 0.7159
69 0.7081
70 0.7095
71 0.7141
72 0.7164
73 0.7156
74 0.7161
75 0.7123
76 0.7155
77 0.71
78 0.7102
79 0.7101
80 0.7138
81 0.7166
82 0.7153
83 0.7117
84 0.711
85 0.7128
86 0.714
87 0.7113
88 0.7122
89 0.7151
90 0.7228
91 0.7153
92 0.7203
93 0.714
94 0.7102
95 0.7162
96 0.7177
97 0.7116
98 0.7119
99 0.7099
100 0.7144
101 0.7082
102 0.716
103 0.7168
104 0.7132
105 0.7109
106 0.7135
107 0.7148
108 0.7165
109 0.7191
110 0.7152
111 0.7082
112 0.7179
113 0.7123
114 0.7159
115 0.7179
116 0.7158
117 0.7153
118 0.7077
119 0.7169
120 0.7108
121 0.715
122 0.7179
123 0.7144
124 0.7163
125 0.7181
126 0.7172
127 0.7086
128 0.7197
129 0.711
130 0.7116
131 0.7153
132 0.7098
133 0.7165
134 0.7147
135 0.711
136 0.7179
137 0.7129
138 0.7176
139 0.7147
140 0.7137
141 0.7193
142 0.7115
143 0.7192
144 0.7167
145 0.7135
146 0.7129
147 0.715
148 0.7183
149 0.7126
150 0.7133
151 0.7097
152 0.7102
153 0.7161
154 0.7108
155 0.7132
156 0.7115
157 0.712
158 0.716
159 0.7199
160 0.7159
161 0.7165
162 0.7118
163 0.7189
164 0.7191
165 0.7192
166 0.7167
167 0.7136
168 0.7158
169 0.7101
170 0.7156
171 0.7193
172 0.7164
173 0.7121
174 0.7146
175 0.7166
176 0.7171
177 0.7183
178 0.711
179 0.7158
180 0.7196
181 0.7154
182 0.7152
183 0.7224
184 0.7179
185 0.7151
186 0.7108
187 0.7199
188 0.713
189 0.7219
190 0.7173
191 0.7157
192 0.7185
193 0.7177
194 0.7176
195 0.7159
196 0.7128
197 0.7156
198 0.7174
199 0.7222
};
% \addlegendentry{Ours(submodularity)}
\addplot [line width=3, forestgreen4416044]
table {%
0 0.0086
1 0.0699
2 0.1319
3 0.1867
4 0.229
5 0.2726
6 0.2934
7 0.303
8 0.3126
9 0.3286
10 0.3436
11 0.3566
12 0.358
13 0.368
14 0.3712
15 0.3846
16 0.3964
17 0.4005
18 0.4111
19 0.4176
20 0.4211
21 0.4338
22 0.4263
23 0.4328
24 0.4383
25 0.4415
26 0.4383
27 0.4514
28 0.4434
29 0.4515
30 0.4503
31 0.4494
32 0.4439
33 0.4493
34 0.4549
35 0.4553
36 0.4572
37 0.4486
38 0.4529
39 0.4473
40 0.4478
41 0.4442
42 0.4497
43 0.4472
44 0.438
45 0.4465
46 0.455
47 0.4536
48 0.4492
49 0.4456
50 0.4445
51 0.4413
52 0.4433
53 0.4514
54 0.4529
55 0.4476
56 0.4435
57 0.4439
58 0.443
59 0.449
60 0.4527
61 0.4418
62 0.4468
63 0.4523
64 0.4402
65 0.4486
66 0.4541
67 0.456
68 0.4509
69 0.4522
70 0.4563
71 0.4508
72 0.451
73 0.4541
74 0.4489
75 0.4479
76 0.4617
77 0.4585
78 0.461
79 0.4578
80 0.4615
81 0.4519
82 0.463
83 0.4619
84 0.4561
85 0.457
86 0.4588
87 0.4524
88 0.4529
89 0.4604
90 0.4558
91 0.4535
92 0.4538
93 0.4555
94 0.4529
95 0.4515
96 0.456
97 0.4496
98 0.4631
99 0.4608
100 0.4533
101 0.45
102 0.4519
103 0.452
104 0.4527
105 0.4586
106 0.4607
107 0.4586
108 0.4514
109 0.4569
110 0.4671
111 0.4633
112 0.4579
113 0.457
114 0.4535
115 0.4627
116 0.4585
117 0.4595
118 0.4632
119 0.4652
120 0.4633
121 0.4615
122 0.4566
123 0.4545
124 0.4581
125 0.4574
126 0.4557
127 0.4562
128 0.4588
129 0.4605
130 0.4605
131 0.4528
132 0.465
133 0.4575
134 0.4671
135 0.4587
136 0.4591
137 0.4664
138 0.4559
139 0.4616
140 0.4617
141 0.461
142 0.4601
143 0.4606
144 0.458
145 0.4589
146 0.4683
147 0.4628
148 0.4582
149 0.4625
150 0.4651
151 0.4598
152 0.4636
153 0.4635
154 0.4666
155 0.4607
156 0.4576
157 0.4543
158 0.4565
159 0.4629
160 0.4638
161 0.4594
162 0.4536
163 0.457
164 0.4592
165 0.4557
166 0.456
167 0.4487
168 0.4556
169 0.4545
170 0.4606
171 0.4569
172 0.4553
173 0.4463
174 0.4536
175 0.4549
176 0.4536
177 0.4523
178 0.4519
179 0.4608
180 0.4526
181 0.4535
182 0.4514
183 0.4585
184 0.4546
185 0.45
186 0.4494
187 0.4577
188 0.4502
189 0.4561
190 0.4494
191 0.4567
192 0.4484
193 0.4572
194 0.4495
195 0.4588
196 0.4503
197 0.4529
198 0.4563
199 0.4579
};
% \addlegendentry{WSGAN}
\end{axis}

\end{tikzpicture}
     % This file was created with tikzplotlib v0.10.1.
\begin{tikzpicture}

\definecolor{darkgray176}{RGB}{176,176,176}
\definecolor{darkorange25512714}{RGB}{255,127,14}
\definecolor{forestgreen4416044}{RGB}{44,160,44}
\definecolor{lightgray204}{RGB}{204,204,204}
\definecolor{steelblue31119180}{RGB}{31,119,180}

\begin{axis}[
legend cell align={left},
legend style={fill opacity=0.8, draw opacity=1, text opacity=1, draw=lightgray204},
tick align=outside,
tick pos=left,
title=\Huge\textbf{AWA2},
x grid style={darkgray176},
xlabel=\Huge{Epoch},
xmajorgrids,
xmin=-9.95, xmax=208.95,
xtick style={color=black},
y grid style={darkgray176},
% ylabel=\Huge{ARI},
ymajorgrids,
ymin=-0.024865, ymax=0.8,
ytick style={color=black},
font=\Huge,
xlabel style={yshift=-14pt},
ylabel style={yshift=14pt}
]
\addplot [line width=3, steelblue31119180]
table {%
0 0.0003
1 0.0785
2 0.0946
3 0.1126
4 0.1202
5 0.1296
6 0.1377
7 0.1387
8 0.1468
9 0.1495
10 0.1564
11 0.1629
12 0.1639
13 0.1651
14 0.164
15 0.1733
16 0.1857
17 0.1817
18 0.1882
19 0.1911
20 0.203
21 0.2236
22 0.2237
23 0.2282
24 0.2418
25 0.2465
26 0.2654
27 0.2705
28 0.2713
29 0.2894
30 0.2916
31 0.3006
32 0.3053
33 0.31
34 0.3134
35 0.3195
36 0.3318
37 0.3242
38 0.3307
39 0.3421
40 0.3325
41 0.3555
42 0.3483
43 0.3493
44 0.3507
45 0.3559
46 0.3509
47 0.3566
48 0.3526
49 0.3573
50 0.3694
51 0.3634
52 0.3644
53 0.3803
54 0.3687
55 0.3706
56 0.3858
57 0.3784
58 0.3804
59 0.3751
60 0.3795
61 0.387
62 0.3836
63 0.3816
64 0.3774
65 0.3997
66 0.3958
67 0.393
68 0.3921
69 0.4035
70 0.3996
71 0.4041
72 0.4014
73 0.4005
74 0.4043
75 0.3952
76 0.4036
77 0.4032
78 0.4112
79 0.4107
80 0.419
81 0.4109
82 0.4066
83 0.4253
84 0.4175
85 0.4183
86 0.4145
87 0.4166
88 0.4198
89 0.427
90 0.4205
91 0.4252
92 0.4224
93 0.4257
94 0.4225
95 0.4264
96 0.4279
97 0.4319
98 0.4329
99 0.4308
100 0.434
101 0.4296
102 0.4302
103 0.4292
104 0.4332
105 0.431
106 0.4329
107 0.4331
108 0.4359
109 0.4449
110 0.4328
111 0.4324
112 0.4412
113 0.4432
114 0.4393
115 0.4484
116 0.4303
117 0.4332
118 0.4394
119 0.437
120 0.4484
121 0.4431
122 0.4397
123 0.4413
124 0.4495
125 0.4521
126 0.4468
127 0.4493
128 0.4456
129 0.4469
130 0.4413
131 0.4522
132 0.4498
133 0.4411
134 0.4525
135 0.4542
136 0.4489
137 0.4339
138 0.4533
139 0.452
140 0.4491
141 0.4409
142 0.4566
143 0.4544
144 0.4472
145 0.4491
146 0.4592
147 0.4537
148 0.4553
149 0.4512
150 0.4572
151 0.4599
152 0.456
153 0.45
154 0.4468
155 0.4544
156 0.4587
157 0.4571
158 0.4626
159 0.4686
160 0.4672
161 0.4569
162 0.4556
163 0.4543
164 0.4603
165 0.4616
166 0.4505
167 0.4615
168 0.4611
169 0.4609
170 0.4569
171 0.4528
172 0.459
173 0.4579
174 0.4621
175 0.4674
176 0.4644
177 0.4663
178 0.4679
179 0.4605
180 0.4633
181 0.4658
182 0.466
183 0.4658
184 0.4657
185 0.4728
186 0.4592
187 0.4636
188 0.4696
189 0.4667
190 0.4603
191 0.4561
192 0.4591
193 0.4584
194 0.4682
195 0.4693
196 0.4574
197 0.4652
198 0.4639
199 0.4784
};
% \addlegendentry{Ours(complete data)}
\addplot [line width=3, darkorange25512714]
table {%
0 -0.0008
1 0.075
2 0.106
3 0.116
4 0.1244
5 0.1396
6 0.1283
7 0.1414
8 0.1506
9 0.1487
10 0.1593
11 0.1538
12 0.17
13 0.1754
14 0.1838
15 0.1974
16 0.2114
17 0.2096
18 0.2054
19 0.2174
20 0.2212
21 0.2289
22 0.2361
23 0.2408
24 0.2545
25 0.2488
26 0.2668
27 0.2634
28 0.2704
29 0.2789
30 0.2737
31 0.282
32 0.2788
33 0.3001
34 0.3009
35 0.3015
36 0.3085
37 0.3107
38 0.317
39 0.3144
40 0.3293
41 0.3247
42 0.3296
43 0.3346
44 0.3341
45 0.3396
46 0.3382
47 0.337
48 0.3414
49 0.344
50 0.3478
51 0.3615
52 0.3567
53 0.359
54 0.3601
55 0.3654
56 0.3575
57 0.3684
58 0.3628
59 0.3663
60 0.3636
61 0.3681
62 0.3685
63 0.3779
64 0.3751
65 0.3802
66 0.376
67 0.3833
68 0.374
69 0.3867
70 0.3903
71 0.3754
72 0.3863
73 0.3962
74 0.3826
75 0.3879
76 0.3922
77 0.3959
78 0.3922
79 0.3938
80 0.3963
81 0.3968
82 0.3923
83 0.4034
84 0.4039
85 0.4003
86 0.3954
87 0.3969
88 0.4088
89 0.4027
90 0.4125
91 0.4078
92 0.3984
93 0.4137
94 0.4143
95 0.4027
96 0.4061
97 0.4163
98 0.4187
99 0.4146
100 0.4167
101 0.4229
102 0.4187
103 0.4111
104 0.4269
105 0.4141
106 0.4241
107 0.424
108 0.4203
109 0.4162
110 0.4228
111 0.4241
112 0.4257
113 0.4213
114 0.4268
115 0.4212
116 0.4402
117 0.4281
118 0.4328
119 0.4342
120 0.4297
121 0.431
122 0.4361
123 0.435
124 0.4335
125 0.4256
126 0.4454
127 0.443
128 0.4397
129 0.4379
130 0.4445
131 0.4432
132 0.4356
133 0.4474
134 0.4372
135 0.4382
136 0.4357
137 0.4449
138 0.4392
139 0.4376
140 0.4499
141 0.4434
142 0.4447
143 0.4475
144 0.4521
145 0.4491
146 0.45
147 0.4371
148 0.4489
149 0.4546
150 0.4477
151 0.4457
152 0.4579
153 0.4552
154 0.4441
155 0.4421
156 0.4477
157 0.4459
158 0.4466
159 0.4505
160 0.4549
161 0.4585
162 0.4419
163 0.4461
164 0.4561
165 0.4567
166 0.4468
167 0.4537
168 0.4598
169 0.46
170 0.4586
171 0.4628
172 0.4516
173 0.4499
174 0.4567
175 0.461
176 0.4573
177 0.4553
178 0.4624
179 0.4573
180 0.464
181 0.4684
182 0.4634
183 0.4592
184 0.4633
185 0.4579
186 0.463
187 0.4582
188 0.4601
189 0.4562
190 0.4725
191 0.4608
192 0.4575
193 0.461
194 0.4576
195 0.4591
196 0.4615
197 0.4675
198 0.4651
199 0.4693
};
% \addlegendentry{Ours(submodularity)}
\addplot [line width=3, forestgreen4416044]
table {%
0 0.0002
1 0.0008
2 -0.0009
3 0.0002
4 0.0354
5 0.0624
6 0.0539
7 0.0557
8 0.063
9 0.0725
10 0.0755
11 0.0871
12 0.0943
13 0.1057
14 0.106
15 0.1114
16 0.109
17 0.1177
18 0.1136
19 0.1168
20 0.1133
21 0.122
22 0.1255
23 0.1288
24 0.139
25 0.1491
26 0.1537
27 0.1587
28 0.1646
29 0.1703
30 0.1702
31 0.1767
32 0.1802
33 0.1807
34 0.1863
35 0.1938
36 0.1905
37 0.1931
38 0.1931
39 0.1974
40 0.1876
41 0.2005
42 0.2067
43 0.2062
44 0.2105
45 0.2107
46 0.2104
47 0.2121
48 0.2204
49 0.2233
50 0.227
51 0.2324
52 0.2312
53 0.2378
54 0.2309
55 0.2406
56 0.2394
57 0.2379
58 0.2435
59 0.2369
60 0.2506
61 0.2495
62 0.2529
63 0.2558
64 0.2514
65 0.2527
66 0.2481
67 0.2509
68 0.2478
69 0.2569
70 0.2623
71 0.2628
72 0.2631
73 0.2693
74 0.2642
75 0.2631
76 0.2665
77 0.2617
78 0.2718
79 0.2634
80 0.2677
81 0.268
82 0.267
83 0.2702
84 0.2703
85 0.2671
86 0.2756
87 0.2719
88 0.2757
89 0.2733
90 0.2761
91 0.269
92 0.2781
93 0.2716
94 0.2659
95 0.2785
96 0.2823
97 0.2875
98 0.2817
99 0.2809
100 0.2771
101 0.2813
102 0.2778
103 0.2785
104 0.2784
105 0.2823
106 0.2743
107 0.2817
108 0.2923
109 0.2875
110 0.2874
111 0.2844
112 0.2861
113 0.2953
114 0.2881
115 0.2906
116 0.2913
117 0.2934
118 0.2951
119 0.2938
120 0.294
121 0.299
122 0.297
123 0.3024
124 0.2926
125 0.2989
126 0.3021
127 0.3019
128 0.302
129 0.307
130 0.3032
131 0.3025
132 0.3018
133 0.3059
134 0.3113
135 0.3149
136 0.3109
137 0.3106
138 0.3044
139 0.3076
140 0.308
141 0.3078
142 0.3193
143 0.3089
144 0.3116
145 0.3117
146 0.314
147 0.3081
148 0.3147
149 0.3194
150 0.3084
151 0.3136
152 0.3102
153 0.3149
154 0.3172
155 0.3157
156 0.3221
157 0.3246
158 0.3187
159 0.3194
160 0.3173
161 0.317
162 0.3121
163 0.3326
164 0.3178
165 0.3286
166 0.3258
167 0.3194
168 0.32
169 0.3294
170 0.3213
171 0.3195
172 0.3267
173 0.3212
174 0.3203
175 0.3242
176 0.3235
177 0.3227
178 0.3268
179 0.3174
180 0.3307
181 0.3214
182 0.3278
183 0.3233
184 0.3349
185 0.33
186 0.3296
187 0.3403
188 0.3362
189 0.3352
190 0.3183
191 0.3346
192 0.3455
193 0.3358
194 0.327
195 0.3329
196 0.3316
197 0.3339
198 0.3358
199 0.3305
};
% \addlegendentry{WSGAN}
\end{axis}

\end{tikzpicture} 
    % This file was created with tikzplotlib v0.10.1.
\begin{tikzpicture}

\definecolor{darkgray176}{RGB}{176,176,176}
\definecolor{darkorange25512714}{RGB}{255,127,14}
\definecolor{forestgreen4416044}{RGB}{44,160,44}
\definecolor{lightgray204}{RGB}{204,204,204}
\definecolor{steelblue31119180}{RGB}{31,119,180}

\begin{axis}[
legend cell align={left},
legend style={fill opacity=0.8, draw opacity=1, text opacity=1, draw=lightgray204},
tick align=outside,
tick pos=left,
title=\Huge\textbf{DomainNet},
x grid style={darkgray176},
xlabel=\Huge{Epoch},
xmajorgrids,
xmin=-9.95, xmax=208.95,
xtick style={color=black},
y grid style={darkgray176},
% ylabel=\Huge{ARI},
ymajorgrids,
ymin=-0.0205, ymax=0.8,
ytick style={color=black},
font=\Huge,
xlabel style={yshift=-14pt},
ylabel style={yshift=14pt}
]
\addplot [line width=3, steelblue31119180]
table {%
0 -0.0001
1 0.0491
2 0.096
3 0.107
4 0.1161
5 0.1328
6 0.141
7 0.1449
8 0.144
9 0.1475
10 0.1464
11 0.1467
12 0.1547
13 0.1535
14 0.1598
15 0.1615
16 0.1714
17 0.1698
18 0.1743
19 0.1799
20 0.1844
21 0.192
22 0.1906
23 0.1961
24 0.2004
25 0.2099
26 0.2145
27 0.2188
28 0.2238
29 0.2227
30 0.2288
31 0.2322
32 0.2372
33 0.2443
34 0.2453
35 0.2488
36 0.2508
37 0.2616
38 0.2635
39 0.2673
40 0.2766
41 0.2757
42 0.2801
43 0.2811
44 0.287
45 0.2852
46 0.2899
47 0.289
48 0.2999
49 0.2929
50 0.3008
51 0.2967
52 0.2999
53 0.3085
54 0.3122
55 0.3078
56 0.3129
57 0.3178
58 0.3157
59 0.3155
60 0.3138
61 0.3199
62 0.3171
63 0.3271
64 0.323
65 0.3276
66 0.3277
67 0.3251
68 0.3325
69 0.3347
70 0.329
71 0.3341
72 0.3355
73 0.337
74 0.3472
75 0.3371
76 0.3399
77 0.3395
78 0.3425
79 0.3529
80 0.344
81 0.3474
82 0.3425
83 0.3477
84 0.3451
85 0.3485
86 0.3447
87 0.3511
88 0.3536
89 0.3572
90 0.3515
91 0.3552
92 0.3542
93 0.3544
94 0.3611
95 0.3611
96 0.3581
97 0.3635
98 0.3612
99 0.3576
100 0.36
101 0.363
102 0.3662
103 0.3655
104 0.366
105 0.3671
106 0.3713
107 0.3681
108 0.3667
109 0.3749
110 0.3716
111 0.3715
112 0.3681
113 0.3766
114 0.3696
115 0.3718
116 0.3742
117 0.3773
118 0.381
119 0.3789
120 0.3705
121 0.3675
122 0.3756
123 0.381
124 0.3744
125 0.3754
126 0.3726
127 0.374
128 0.3836
129 0.3727
130 0.3721
131 0.3777
132 0.3755
133 0.3844
134 0.3772
135 0.3864
136 0.3795
137 0.3816
138 0.3884
139 0.3856
140 0.378
141 0.3702
142 0.3803
143 0.3824
144 0.3828
145 0.3841
146 0.385
147 0.3889
148 0.3861
149 0.394
150 0.3794
151 0.3857
152 0.3855
153 0.3865
154 0.3974
155 0.3808
156 0.3766
157 0.3871
158 0.3909
159 0.3845
160 0.3916
161 0.3868
162 0.3802
163 0.3926
164 0.3959
165 0.3811
166 0.3884
167 0.3864
168 0.3937
169 0.3851
170 0.397
171 0.3926
172 0.3941
173 0.3909
174 0.396
175 0.3939
176 0.3961
177 0.3942
178 0.3988
179 0.3958
180 0.3922
181 0.3936
182 0.3918
183 0.3954
184 0.3993
185 0.3965
186 0.3941
187 0.3947
188 0.3993
189 0.3968
190 0.394
191 0.3963
192 0.3989
193 0.4014
194 0.3907
195 0.3965
196 0.3986
197 0.4005
198 0.4007
199 0.4079
};
% \addlegendentry{Ours(complete data)}
\addplot [line width=3, darkorange25512714]
table {%
0 0.001
1 0.0433
2 0.0771
3 0.1014
4 0.1284
5 0.1349
6 0.1424
7 0.1437
8 0.1404
9 0.1469
10 0.1439
11 0.1437
12 0.1458
13 0.1458
14 0.1648
15 0.1463
16 0.1626
17 0.162
18 0.1598
19 0.1667
20 0.1705
21 0.1773
22 0.1858
23 0.1876
24 0.1983
25 0.1977
26 0.2044
27 0.2105
28 0.2226
29 0.2157
30 0.2231
31 0.2296
32 0.2357
33 0.237
34 0.2406
35 0.244
36 0.2504
37 0.2534
38 0.2553
39 0.2588
40 0.2671
41 0.2672
42 0.2673
43 0.2768
44 0.2722
45 0.2777
46 0.2788
47 0.2781
48 0.2829
49 0.2862
50 0.2935
51 0.2877
52 0.2941
53 0.2858
54 0.2973
55 0.2982
56 0.3006
57 0.3013
58 0.2999
59 0.307
60 0.3107
61 0.3088
62 0.311
63 0.3152
64 0.3188
65 0.3141
66 0.3174
67 0.3212
68 0.3247
69 0.3222
70 0.3245
71 0.3209
72 0.3376
73 0.3342
74 0.3258
75 0.3377
76 0.3371
77 0.3315
78 0.336
79 0.3367
80 0.3273
81 0.3276
82 0.3406
83 0.3396
84 0.3335
85 0.3402
86 0.3353
87 0.344
88 0.3466
89 0.3515
90 0.3526
91 0.3467
92 0.3496
93 0.3481
94 0.3427
95 0.354
96 0.3562
97 0.353
98 0.3501
99 0.3586
100 0.3577
101 0.3524
102 0.3538
103 0.3607
104 0.3575
105 0.363
106 0.3571
107 0.3612
108 0.3625
109 0.3574
110 0.3579
111 0.3651
112 0.3598
113 0.3644
114 0.3624
115 0.3617
116 0.3607
117 0.3636
118 0.3668
119 0.3643
120 0.3699
121 0.3648
122 0.3674
123 0.3669
124 0.3696
125 0.3714
126 0.3775
127 0.3755
128 0.3756
129 0.3724
130 0.3736
131 0.3774
132 0.373
133 0.3664
134 0.3736
135 0.3697
136 0.3718
137 0.3734
138 0.3765
139 0.3828
140 0.3784
141 0.3818
142 0.376
143 0.3826
144 0.3748
145 0.3836
146 0.3815
147 0.3793
148 0.3812
149 0.3818
150 0.3811
151 0.3832
152 0.3802
153 0.3831
154 0.3797
155 0.3833
156 0.3804
157 0.3783
158 0.384
159 0.3888
160 0.3817
161 0.3862
162 0.3842
163 0.3886
164 0.3856
165 0.3931
166 0.3884
167 0.3868
168 0.395
169 0.3868
170 0.3817
171 0.3937
172 0.382
173 0.3857
174 0.382
175 0.3925
176 0.392
177 0.3895
178 0.3902
179 0.3882
180 0.3901
181 0.396
182 0.3933
183 0.3974
184 0.3871
185 0.3958
186 0.3951
187 0.3899
188 0.3979
189 0.3992
190 0.3944
191 0.3949
192 0.3911
193 0.4046
194 0.4006
195 0.4
196 0.4016
197 0.4
198 0.3899
199 0.4058
};
% \addlegendentry{Ours(submodularity)}
\addplot [line width=3, forestgreen4416044]
table {%
0 0.0007
1 0.0231
2 0.0576
3 0.0688
4 0.0813
5 0.0951
6 0.095
7 0.1045
8 0.1075
9 0.113
10 0.1223
11 0.1296
12 0.133
13 0.1399
14 0.143
15 0.1548
16 0.1543
17 0.1557
18 0.1667
19 0.1725
20 0.1674
21 0.1666
22 0.1708
23 0.1807
24 0.1818
25 0.1758
26 0.1861
27 0.177
28 0.1869
29 0.1874
30 0.1894
31 0.1889
32 0.1965
33 0.1986
34 0.1946
35 0.1977
36 0.2069
37 0.2048
38 0.2083
39 0.207
40 0.2092
41 0.2103
42 0.2099
43 0.2143
44 0.2224
45 0.2275
46 0.2154
47 0.2218
48 0.2251
49 0.2215
50 0.226
51 0.2271
52 0.2334
53 0.2291
54 0.2266
55 0.2297
56 0.2273
57 0.234
58 0.2374
59 0.2333
60 0.2296
61 0.2359
62 0.2386
63 0.2395
64 0.2388
65 0.2374
66 0.2368
67 0.2363
68 0.2415
69 0.2428
70 0.2448
71 0.2474
72 0.2395
73 0.2479
74 0.2497
75 0.2454
76 0.2436
77 0.2441
78 0.2493
79 0.2507
80 0.251
81 0.2558
82 0.2512
83 0.2532
84 0.2493
85 0.2537
86 0.2533
87 0.2545
88 0.2601
89 0.2621
90 0.2625
91 0.2689
92 0.2608
93 0.258
94 0.261
95 0.2653
96 0.2614
97 0.2645
98 0.2628
99 0.2652
100 0.2655
101 0.272
102 0.2642
103 0.2625
104 0.2681
105 0.2709
106 0.2631
107 0.2715
108 0.2758
109 0.2737
110 0.2793
111 0.2772
112 0.272
113 0.266
114 0.2674
115 0.2783
116 0.2736
117 0.2791
118 0.2741
119 0.2777
120 0.2695
121 0.2687
122 0.2775
123 0.2833
124 0.2847
125 0.2848
126 0.2792
127 0.2775
128 0.284
129 0.2833
130 0.281
131 0.2837
132 0.2869
133 0.2873
134 0.287
135 0.287
136 0.2829
137 0.2867
138 0.288
139 0.2848
140 0.2893
141 0.2925
142 0.2918
143 0.2941
144 0.2977
145 0.2916
146 0.2927
147 0.2946
148 0.2927
149 0.298
150 0.2999
151 0.3031
152 0.3013
153 0.2971
154 0.2942
155 0.3014
156 0.3009
157 0.3057
158 0.3055
159 0.3051
160 0.3087
161 0.3061
162 0.2977
163 0.3067
164 0.3126
165 0.3054
166 0.3095
167 0.3088
168 0.3106
169 0.307
170 0.3104
171 0.3075
172 0.309
173 0.3057
174 0.3171
175 0.3106
176 0.3091
177 0.3119
178 0.311
179 0.3045
180 0.3094
181 0.3057
182 0.3133
183 0.3106
184 0.315
185 0.3229
186 0.3179
187 0.3161
188 0.3249
189 0.3261
190 0.3175
191 0.3178
192 0.3186
193 0.322
194 0.3207
195 0.3143
196 0.311
197 0.3223
198 0.3184
199 0.3237
};
% \addlegendentry{WSGAN}
\end{axis}

\end{tikzpicture}
    % This file was created with tikzplotlib v0.10.1.
\begin{tikzpicture}

\definecolor{darkgray176}{RGB}{176,176,176}
\definecolor{darkorange25512714}{RGB}{255,127,14}
\definecolor{forestgreen4416044}{RGB}{44,160,44}
\definecolor{lightgray204}{RGB}{204,204,204}
\definecolor{steelblue31119180}{RGB}{31,119,180}

\begin{axis}[
legend cell align={left},
legend style={
  fill opacity=0.8,
  draw opacity=1,
  text opacity=1,
  at={(0.97,0.03)},
  anchor=south east,
  draw=lightgray204
},
tick align=outside,
tick pos=left,
title=\Huge\textbf{MNIST},
x grid style={darkgray176},
xlabel=\Huge{Epoch},
xmajorgrids,
xmin=-4.95, xmax=103.95,
xtick style={color=black},
y grid style={darkgray176},
% ylabel=\Huge{ARI},
ymajorgrids,
ymin=-0.042145, ymax=0.893845,
ytick style={color=black},
font=\Huge,
xlabel style={yshift=-14pt},
ylabel style={yshift=14pt}
]
\addplot [line width=3, steelblue31119180]
table {%
0 0.0004
1 0.4134
2 0.7518
3 0.7663
4 0.7845
5 0.7853
6 0.7891
7 0.793
8 0.7901
9 0.7911
10 0.7913
11 0.792
12 0.7887
13 0.7929
14 0.791
15 0.7856
16 0.7871
17 0.7863
18 0.7859
19 0.7858
20 0.7869
21 0.7862
22 0.7948
23 0.7862
24 0.7834
25 0.7878
26 0.7893
27 0.7938
28 0.7895
29 0.7889
30 0.7851
31 0.7877
32 0.7848
33 0.7885
34 0.7893
35 0.7901
36 0.7917
37 0.7856
38 0.7853
39 0.7871
40 0.7897
41 0.7852
42 0.7883
43 0.7866
44 0.7883
45 0.7894
46 0.7878
47 0.7888
48 0.7877
49 0.7823
50 0.7828
51 0.7898
52 0.7874
53 0.7841
54 0.7861
55 0.7823
56 0.7885
57 0.7888
58 0.7846
59 0.7892
60 0.7887
61 0.7838
62 0.7869
63 0.7893
64 0.7863
65 0.787
66 0.7858
67 0.7883
68 0.7875
69 0.7851
70 0.7892
71 0.7897
72 0.7849
73 0.7855
74 0.7841
75 0.78
76 0.7841
77 0.7815
78 0.7856
79 0.7835
80 0.7865
81 0.7881
82 0.782
83 0.7837
84 0.7825
85 0.7897
86 0.781
87 0.7847
88 0.7816
89 0.7847
90 0.7828
91 0.7849
92 0.7839
93 0.7822
94 0.7858
95 0.7838
96 0.7842
97 0.7831
98 0.781
99 0.7814
};
% \addlegendentry{Ours(complete data)}
\addplot [line width=3, darkorange25512714]
table {%
0 0.0028
1 0.6066
2 0.7646
3 0.7811
4 0.7989
5 0.808
6 0.8106
7 0.8078
8 0.8041
9 0.7996
10 0.8051
11 0.8028
12 0.8035
13 0.7993
14 0.8033
15 0.7997
16 0.8006
17 0.809
18 0.8061
19 0.8097
20 0.81
21 0.8099
22 0.8145
23 0.814
24 0.8108
25 0.8123
26 0.8124
27 0.8132
28 0.815
29 0.8156
30 0.8155
31 0.8145
32 0.8164
33 0.8173
34 0.8209
35 0.8168
36 0.8158
37 0.8153
38 0.8153
39 0.8178
40 0.8189
41 0.8135
42 0.8141
43 0.8143
44 0.8135
45 0.8156
46 0.8169
47 0.8149
48 0.8172
49 0.8216
50 0.8194
51 0.8138
52 0.8187
53 0.816
54 0.8175
55 0.815
56 0.8169
57 0.8194
58 0.8161
59 0.82
60 0.8169
61 0.8185
62 0.8165
63 0.8176
64 0.8187
65 0.8204
66 0.8188
67 0.8189
68 0.8173
69 0.8193
70 0.8156
71 0.8206
72 0.8169
73 0.8215
74 0.818
75 0.8213
76 0.8214
77 0.8177
78 0.8172
79 0.8155
80 0.8143
81 0.8198
82 0.8217
83 0.8207
84 0.8198
85 0.8123
86 0.8236
87 0.8137
88 0.8231
89 0.8194
90 0.8184
91 0.8203
92 0.8182
93 0.8213
94 0.8187
95 0.8243
96 0.8213
97 0.8204
98 0.8225
99 0.8175
};
% \addlegendentry{Ours(submodularity)}
\addplot [line width=3, forestgreen4416044]
table {%
0 0.1527
1 0.5519
2 0.7013
3 0.7866
4 0.8249
5 0.8464
6 0.8513
7 0.8493
8 0.8462
9 0.8438
10 0.8426
11 0.8387
12 0.8331
13 0.8322
14 0.8239
15 0.8208
16 0.8185
17 0.8185
18 0.8167
19 0.8127
20 0.8142
21 0.8121
22 0.8091
23 0.8095
24 0.8106
25 0.804
26 0.8086
27 0.8081
28 0.8024
29 0.8101
30 0.803
31 0.8083
32 0.8103
33 0.8134
34 0.8103
35 0.8077
36 0.8129
37 0.8137
38 0.8083
39 0.8176
40 0.8133
41 0.811
42 0.8177
43 0.8163
44 0.8119
45 0.8145
46 0.8189
47 0.8165
48 0.8212
49 0.8167
50 0.8104
51 0.8183
52 0.8217
53 0.814
54 0.8163
55 0.8178
56 0.8144
57 0.8131
58 0.8147
59 0.8161
60 0.8185
61 0.8195
62 0.8232
63 0.8211
64 0.8128
65 0.8156
66 0.8155
67 0.8139
68 0.8032
69 0.8172
70 0.8144
71 0.802
72 0.7974
73 0.8096
74 0.8055
75 0.8024
76 0.8031
77 0.8036
78 0.7994
79 0.7934
80 0.7965
81 0.8004
82 0.7999
83 0.7988
84 0.8035
85 0.8013
86 0.8009
87 0.7906
88 0.7972
89 0.8016
90 0.797
91 0.7962
92 0.7907
93 0.7876
94 0.7937
95 0.792
96 0.7868
97 0.7917
98 0.7893
99 0.7909
};
% \addlegendentry{WSGAN}
\end{axis}

\end{tikzpicture}
    
}
\end{figure}
\vspace{-22pt}
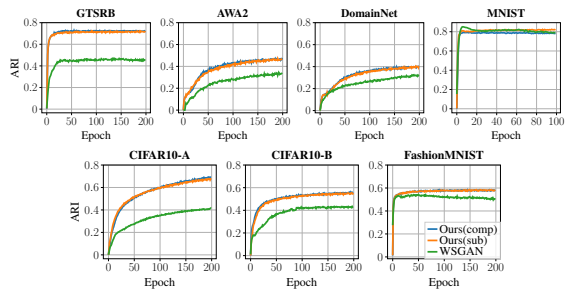
\begin{figure}[ht!]
    \centering
     \resizebox{0.33\textwidth}{!}
 {
 
    % This file was created with tikzplotlib v0.10.1.
\begin{tikzpicture}

\definecolor{darkgray176}{RGB}{176,176,176}
\definecolor{darkorange25512714}{RGB}{255,127,14}
\definecolor{forestgreen4416044}{RGB}{44,160,44}
\definecolor{lightgray204}{RGB}{204,204,204}
\definecolor{steelblue31119180}{RGB}{31,119,180}

\begin{axis}[
legend cell align={left},
legend style={
  fill opacity=0.8,
  draw opacity=1,
  text opacity=1,
  at={(0.03,0.97)},
  anchor=north west,
  draw=lightgray204
},
tick align=outside,
tick pos=left,
title=\Huge\textbf{CIFAR10-A},
x grid style={darkgray176},
xlabel=\Huge{Epoch},
xmajorgrids,
xmin=-9.95, xmax=208.95,
xtick style={color=black},
y grid style={darkgray176},
ylabel=\Huge{ARI},
ymajorgrids,
ymin=-0.034635, ymax=0.8,
ytick style={color=black},
font=\Huge,
xlabel style={yshift=-14pt},
ylabel style={yshift=14pt}
]
\addplot [line width=3, steelblue31119180]
table {%
0 0
1 0.0482
2 0.0646
3 0.0974
4 0.1338
5 0.1667
6 0.1849
7 0.2088
8 0.219
9 0.2391
10 0.2485
11 0.2682
12 0.2808
13 0.302
14 0.3112
15 0.3275
16 0.3406
17 0.3471
18 0.3558
19 0.3691
20 0.3745
21 0.3839
22 0.392
23 0.4013
24 0.4027
25 0.4112
26 0.415
27 0.4242
28 0.4305
29 0.4326
30 0.4372
31 0.4451
32 0.4479
33 0.4494
34 0.4556
35 0.4604
36 0.4621
37 0.4629
38 0.4703
39 0.4722
40 0.4765
41 0.4794
42 0.4839
43 0.4899
44 0.4856
45 0.4876
46 0.4991
47 0.4966
48 0.503
49 0.5081
50 0.5077
51 0.5161
52 0.5124
53 0.5168
54 0.5139
55 0.5246
56 0.5255
57 0.5214
58 0.5236
59 0.5317
60 0.5316
61 0.5343
62 0.5318
63 0.541
64 0.5419
65 0.5383
66 0.551
67 0.5521
68 0.5494
69 0.5488
70 0.554
71 0.5515
72 0.5587
73 0.5569
74 0.5579
75 0.5638
76 0.5658
77 0.5712
78 0.569
79 0.5714
80 0.571
81 0.5759
82 0.5718
83 0.5812
84 0.5758
85 0.5819
86 0.5796
87 0.5771
88 0.5866
89 0.5839
90 0.5897
91 0.5853
92 0.5901
93 0.5893
94 0.5935
95 0.592
96 0.5975
97 0.5919
98 0.5945
99 0.5961
100 0.5992
101 0.6
102 0.6043
103 0.6007
104 0.6107
105 0.6081
106 0.6028
107 0.6128
108 0.6061
109 0.6109
110 0.6134
111 0.6155
112 0.615
113 0.6181
114 0.6229
115 0.6228
116 0.6191
117 0.6218
118 0.6222
119 0.6262
120 0.6239
121 0.6272
122 0.6295
123 0.6274
124 0.6249
125 0.6314
126 0.632
127 0.6304
128 0.6343
129 0.632
130 0.6388
131 0.638
132 0.6337
133 0.6336
134 0.6439
135 0.6413
136 0.6417
137 0.64
138 0.6421
139 0.6518
140 0.6439
141 0.6534
142 0.6459
143 0.6489
144 0.6468
145 0.6517
146 0.6475
147 0.6551
148 0.65
149 0.6522
150 0.6601
151 0.6574
152 0.6545
153 0.658
154 0.6573
155 0.6606
156 0.6618
157 0.6636
158 0.6613
159 0.6585
160 0.6592
161 0.6621
162 0.6612
163 0.6702
164 0.6613
165 0.6678
166 0.6641
167 0.6758
168 0.6694
169 0.6709
170 0.6704
171 0.6708
172 0.6737
173 0.6744
174 0.676
175 0.6752
176 0.6778
177 0.6756
178 0.6749
179 0.6769
180 0.6818
181 0.6827
182 0.6818
183 0.6778
184 0.686
185 0.6784
186 0.6883
187 0.6861
188 0.6821
189 0.6848
190 0.6858
191 0.6859
192 0.6894
193 0.6886
194 0.6927
195 0.6901
196 0.6909
197 0.6869
198 0.6908
199 0.69
};
% \addlegendentry{Ours(complete data)}
\addplot [line width=3, darkorange25512714]
table {%
0 0.0006
1 0.0389
2 0.1106
3 0.1279
4 0.155
5 0.1861
6 0.2031
7 0.2284
8 0.2552
9 0.2741
10 0.289
11 0.3028
12 0.3192
13 0.33
14 0.3425
15 0.3518
16 0.3623
17 0.3735
18 0.3857
19 0.3918
20 0.3932
21 0.4041
22 0.4132
23 0.4191
24 0.4191
25 0.4269
26 0.4314
27 0.4404
28 0.4424
29 0.4486
30 0.4572
31 0.4509
32 0.4605
33 0.4655
34 0.4695
35 0.4652
36 0.4746
37 0.4792
38 0.4835
39 0.4845
40 0.4829
41 0.4882
42 0.4909
43 0.4982
44 0.5011
45 0.5064
46 0.509
47 0.5131
48 0.5047
49 0.5122
50 0.5148
51 0.5173
52 0.5217
53 0.5207
54 0.5273
55 0.5279
56 0.5234
57 0.5318
58 0.5341
59 0.5344
60 0.5353
61 0.5396
62 0.5379
63 0.5426
64 0.542
65 0.5414
66 0.5472
67 0.548
68 0.5546
69 0.553
70 0.5511
71 0.5573
72 0.5614
73 0.5613
74 0.5629
75 0.5628
76 0.5673
77 0.5666
78 0.5682
79 0.5658
80 0.5714
81 0.57
82 0.5734
83 0.5746
84 0.5745
85 0.5809
86 0.5801
87 0.5739
88 0.5777
89 0.5806
90 0.5842
91 0.5862
92 0.5858
93 0.5847
94 0.5891
95 0.594
96 0.5918
97 0.5968
98 0.5973
99 0.5922
100 0.5961
101 0.5943
102 0.5968
103 0.6042
104 0.603
105 0.6018
106 0.6011
107 0.5991
108 0.6064
109 0.6056
110 0.6091
111 0.606
112 0.6186
113 0.611
114 0.6073
115 0.6164
116 0.6143
117 0.6092
118 0.6189
119 0.615
120 0.622
121 0.6196
122 0.6169
123 0.6216
124 0.6155
125 0.6231
126 0.6206
127 0.6296
128 0.6261
129 0.6277
130 0.6254
131 0.6319
132 0.6263
133 0.6272
134 0.6279
135 0.6294
136 0.6244
137 0.6308
138 0.6311
139 0.6323
140 0.6403
141 0.6374
142 0.6373
143 0.6331
144 0.6412
145 0.6423
146 0.6443
147 0.6451
148 0.6431
149 0.6425
150 0.6478
151 0.6428
152 0.6482
153 0.6438
154 0.6477
155 0.6463
156 0.6499
157 0.6519
158 0.6499
159 0.6534
160 0.6539
161 0.6547
162 0.6563
163 0.6491
164 0.6611
165 0.6548
166 0.6532
167 0.6557
168 0.6602
169 0.6551
170 0.6648
171 0.6594
172 0.662
173 0.6614
174 0.6586
175 0.6667
176 0.66
177 0.6609
178 0.6611
179 0.6685
180 0.6687
181 0.6623
182 0.6691
183 0.6617
184 0.6693
185 0.6701
186 0.6682
187 0.6695
188 0.6733
189 0.6693
190 0.6722
191 0.665
192 0.6716
193 0.6764
194 0.6726
195 0.6772
196 0.6778
197 0.6664
198 0.6749
199 0.6811
};
% \addlegendentry{Ours(submodularity)}
\addplot [line width=3, forestgreen4416044]
table {%
0 0.0013
1 0.0231
2 0.0451
3 0.0682
4 0.0779
5 0.0832
6 0.0949
7 0.1115
8 0.1225
9 0.1358
10 0.1432
11 0.1574
12 0.1665
13 0.1775
14 0.1848
15 0.1868
16 0.1911
17 0.1946
18 0.1983
19 0.2041
20 0.2038
21 0.2066
22 0.2114
23 0.2134
24 0.216
25 0.2177
26 0.2213
27 0.2215
28 0.2271
29 0.2284
30 0.2304
31 0.2271
32 0.234
33 0.2362
34 0.2447
35 0.2435
36 0.2466
37 0.249
38 0.2514
39 0.2568
40 0.2585
41 0.2568
42 0.2611
43 0.2658
44 0.266
45 0.2683
46 0.2695
47 0.2702
48 0.2712
49 0.2715
50 0.2764
51 0.2754
52 0.2829
53 0.2842
54 0.2853
55 0.2892
56 0.2886
57 0.2887
58 0.2936
59 0.2947
60 0.2976
61 0.2966
62 0.3012
63 0.3069
64 0.3033
65 0.3065
66 0.3064
67 0.3119
68 0.3079
69 0.3157
70 0.3148
71 0.3178
72 0.3119
73 0.3195
74 0.3262
75 0.3224
76 0.3208
77 0.3271
78 0.327
79 0.3329
80 0.3303
81 0.3334
82 0.3298
83 0.3336
84 0.3344
85 0.3373
86 0.3365
87 0.3387
88 0.3358
89 0.3413
90 0.3414
91 0.3456
92 0.346
93 0.3434
94 0.3488
95 0.3511
96 0.3454
97 0.3451
98 0.3486
99 0.351
100 0.349
101 0.3506
102 0.3521
103 0.3585
104 0.3564
105 0.3547
106 0.3553
107 0.3552
108 0.3563
109 0.36
110 0.363
111 0.3578
112 0.3635
113 0.3648
114 0.3637
115 0.3646
116 0.3653
117 0.3701
118 0.367
119 0.3643
120 0.3655
121 0.3692
122 0.3665
123 0.3686
124 0.374
125 0.3715
126 0.3699
127 0.37
128 0.3742
129 0.3793
130 0.3781
131 0.3768
132 0.3778
133 0.3796
134 0.3776
135 0.382
136 0.3796
137 0.378
138 0.3828
139 0.3815
140 0.3827
141 0.3878
142 0.3828
143 0.3875
144 0.384
145 0.3868
146 0.3851
147 0.3879
148 0.3913
149 0.3853
150 0.3904
151 0.3908
152 0.3886
153 0.3874
154 0.3919
155 0.3928
156 0.3943
157 0.3912
158 0.3921
159 0.3945
160 0.396
161 0.3922
162 0.399
163 0.3945
164 0.3961
165 0.3942
166 0.3988
167 0.3994
168 0.3971
169 0.3992
170 0.3973
171 0.4002
172 0.4006
173 0.3994
174 0.4004
175 0.4001
176 0.3989
177 0.4052
178 0.403
179 0.4032
180 0.4038
181 0.4015
182 0.407
183 0.408
184 0.4091
185 0.4067
186 0.4023
187 0.4047
188 0.4031
189 0.405
190 0.4069
191 0.4049
192 0.4073
193 0.4052
194 0.4087
195 0.4084
196 0.4073
197 0.4129
198 0.4163
199 0.4134
};
% \addlegendentry{WSGAN}
\end{axis}

\end{tikzpicture}    
    % This file was created with tikzplotlib v0.10.1.
\begin{tikzpicture}

\definecolor{darkgray176}{RGB}{176,176,176}
\definecolor{darkorange25512714}{RGB}{255,127,14}
\definecolor{forestgreen4416044}{RGB}{44,160,44}
\definecolor{lightgray204}{RGB}{204,204,204}
\definecolor{steelblue31119180}{RGB}{31,119,180}

\begin{axis}[
legend cell align={left},
legend style={fill opacity=0.8, draw opacity=1, text opacity=1, draw=lightgray204},
tick align=outside,
tick pos=left,
title=\Huge\textbf{CIFAR10-B},
x grid style={darkgray176},
xlabel=\Huge{Epoch},
xmajorgrids,
xmin=-9.95, xmax=208.95,
xtick style={color=black},
y grid style={darkgray176},
% ylabel=\Huge{ARI},
ymajorgrids,
ymin=-0.02785, ymax=0.8,
ytick style={color=black},
font=\Huge,
xlabel style={yshift=-14pt},
ylabel style={yshift=14pt}
]
\addplot [line width=3, steelblue31119180]
table {%
0 0.0002
1 0.1344
2 0.2173
3 0.2457
4 0.265
5 0.2734
6 0.293
7 0.3123
8 0.3372
9 0.3532
10 0.3694
11 0.3801
12 0.3866
13 0.3979
14 0.4088
15 0.4171
16 0.4148
17 0.4264
18 0.4306
19 0.4363
20 0.44
21 0.4468
22 0.4501
23 0.4555
24 0.4555
25 0.4575
26 0.4661
27 0.4622
28 0.4627
29 0.4734
30 0.4743
31 0.4806
32 0.4769
33 0.4769
34 0.479
35 0.4788
36 0.4779
37 0.4804
38 0.487
39 0.4858
40 0.4839
41 0.4891
42 0.4938
43 0.4943
44 0.4944
45 0.4958
46 0.4974
47 0.5014
48 0.4992
49 0.5005
50 0.4993
51 0.4946
52 0.5027
53 0.5032
54 0.5067
55 0.5027
56 0.5048
57 0.51
58 0.5101
59 0.5055
60 0.5139
61 0.5119
62 0.5128
63 0.5147
64 0.512
65 0.5133
66 0.5131
67 0.514
68 0.5187
69 0.525
70 0.5176
71 0.5167
72 0.5186
73 0.5154
74 0.5183
75 0.5256
76 0.5211
77 0.5249
78 0.522
79 0.5226
80 0.5222
81 0.5195
82 0.5261
83 0.5212
84 0.5256
85 0.5254
86 0.5281
87 0.5238
88 0.5262
89 0.5332
90 0.5267
91 0.5257
92 0.5293
93 0.538
94 0.5308
95 0.5282
96 0.5311
97 0.5298
98 0.5373
99 0.534
100 0.5339
101 0.5309
102 0.5309
103 0.5275
104 0.5312
105 0.5331
106 0.5317
107 0.5361
108 0.5322
109 0.5323
110 0.5371
111 0.5313
112 0.5313
113 0.5369
114 0.5374
115 0.5375
116 0.5311
117 0.5393
118 0.5373
119 0.5393
120 0.541
121 0.5378
122 0.5454
123 0.5396
124 0.5381
125 0.5418
126 0.5424
127 0.5426
128 0.5408
129 0.5412
130 0.5437
131 0.5467
132 0.5406
133 0.5489
134 0.5387
135 0.5447
136 0.544
137 0.5402
138 0.5466
139 0.5481
140 0.5393
141 0.5469
142 0.5384
143 0.5508
144 0.5447
145 0.5449
146 0.549
147 0.5468
148 0.5443
149 0.544
150 0.547
151 0.5446
152 0.5504
153 0.548
154 0.5471
155 0.5494
156 0.5497
157 0.5482
158 0.5492
159 0.5496
160 0.5462
161 0.55
162 0.55
163 0.5561
164 0.5559
165 0.5464
166 0.5519
167 0.5488
168 0.5523
169 0.5426
170 0.5524
171 0.552
172 0.5509
173 0.5535
174 0.553
175 0.549
176 0.5526
177 0.5491
178 0.5555
179 0.5554
180 0.5483
181 0.5547
182 0.5534
183 0.5493
184 0.5544
185 0.5504
186 0.55
187 0.5544
188 0.5529
189 0.5538
190 0.5547
191 0.5592
192 0.5495
193 0.5541
194 0.5612
195 0.5513
196 0.5565
197 0.559
198 0.5555
199 0.5542
};
% \addlegendentry{Ours(complete data)}
\addplot [line width=3, darkorange25512714]
table {%
0 0.0006
1 0.0923
2 0.151
3 0.174
4 0.2202
5 0.2535
6 0.278
7 0.2832
8 0.3056
9 0.3171
10 0.3308
11 0.3472
12 0.3525
13 0.369
14 0.3856
15 0.3893
16 0.3946
17 0.4011
18 0.4164
19 0.4255
20 0.4239
21 0.4287
22 0.4335
23 0.4357
24 0.445
25 0.441
26 0.4516
27 0.4545
28 0.4632
29 0.4651
30 0.4621
31 0.4668
32 0.4633
33 0.4701
34 0.475
35 0.4654
36 0.4721
37 0.4733
38 0.4818
39 0.479
40 0.4747
41 0.48
42 0.4855
43 0.4873
44 0.4874
45 0.4808
46 0.4908
47 0.4895
48 0.488
49 0.4898
50 0.4903
51 0.4952
52 0.491
53 0.4995
54 0.4942
55 0.4982
56 0.4963
57 0.5051
58 0.5022
59 0.501
60 0.5029
61 0.5033
62 0.5038
63 0.5043
64 0.5123
65 0.508
66 0.5054
67 0.5062
68 0.511
69 0.5093
70 0.5047
71 0.5128
72 0.5112
73 0.5061
74 0.5081
75 0.5145
76 0.5136
77 0.5077
78 0.5157
79 0.5202
80 0.5152
81 0.5182
82 0.5212
83 0.5179
84 0.5219
85 0.5257
86 0.5219
87 0.5225
88 0.5213
89 0.5227
90 0.5151
91 0.5217
92 0.5224
93 0.5227
94 0.5262
95 0.5248
96 0.5277
97 0.5278
98 0.5235
99 0.5275
100 0.5322
101 0.5281
102 0.5305
103 0.5261
104 0.5289
105 0.5305
106 0.5319
107 0.5307
108 0.5277
109 0.5264
110 0.5333
111 0.5272
112 0.5301
113 0.5339
114 0.534
115 0.5326
116 0.5334
117 0.535
118 0.5349
119 0.5369
120 0.5347
121 0.538
122 0.5331
123 0.5371
124 0.5382
125 0.5356
126 0.5377
127 0.5338
128 0.5356
129 0.5391
130 0.5378
131 0.5356
132 0.5417
133 0.5352
134 0.5401
135 0.5373
136 0.5396
137 0.5407
138 0.5365
139 0.544
140 0.5389
141 0.5452
142 0.544
143 0.5401
144 0.5424
145 0.5433
146 0.5369
147 0.5435
148 0.5379
149 0.537
150 0.5426
151 0.5416
152 0.5395
153 0.5465
154 0.5427
155 0.5338
156 0.5398
157 0.5405
158 0.5443
159 0.5377
160 0.5445
161 0.5465
162 0.5424
163 0.5453
164 0.5455
165 0.5505
166 0.543
167 0.5437
168 0.5464
169 0.5502
170 0.5443
171 0.5399
172 0.5425
173 0.541
174 0.5524
175 0.5484
176 0.5469
177 0.5477
178 0.5441
179 0.5494
180 0.5379
181 0.5448
182 0.548
183 0.5452
184 0.5489
185 0.5457
186 0.5479
187 0.5485
188 0.5478
189 0.5539
190 0.5499
191 0.5477
192 0.5438
193 0.5491
194 0.5465
195 0.5515
196 0.5546
197 0.5477
198 0.545
199 0.5537
};
% \addlegendentry{Ours(submodularity)}
\addplot [line width=3, forestgreen4416044]
table {%
0 0.0113
1 0.1015
2 0.14
3 0.1547
4 0.1683
5 0.1798
6 0.1813
7 0.1771
8 0.1761
9 0.1819
10 0.1883
11 0.2013
12 0.2105
13 0.2122
14 0.2196
15 0.2287
16 0.2354
17 0.2334
18 0.2379
19 0.2468
20 0.2485
21 0.2522
22 0.2541
23 0.2674
24 0.2611
25 0.2667
26 0.2709
27 0.2751
28 0.28
29 0.2842
30 0.2914
31 0.2977
32 0.3002
33 0.3083
34 0.3182
35 0.3239
36 0.3254
37 0.3298
38 0.3371
39 0.3377
40 0.3353
41 0.3469
42 0.3393
43 0.3474
44 0.3432
45 0.3486
46 0.3531
47 0.3494
48 0.3628
49 0.3669
50 0.3612
51 0.3699
52 0.3681
53 0.3666
54 0.3715
55 0.3735
56 0.3755
57 0.3771
58 0.3794
59 0.3771
60 0.3774
61 0.3765
62 0.3824
63 0.3838
64 0.3934
65 0.381
66 0.386
67 0.387
68 0.3903
69 0.3896
70 0.3892
71 0.3948
72 0.3941
73 0.3975
74 0.3959
75 0.3987
76 0.3963
77 0.4053
78 0.4111
79 0.4131
80 0.4135
81 0.4083
82 0.398
83 0.3997
84 0.3978
85 0.4014
86 0.401
87 0.4021
88 0.4053
89 0.4227
90 0.4221
91 0.4211
92 0.4244
93 0.4205
94 0.4149
95 0.4199
96 0.4167
97 0.4197
98 0.4279
99 0.4182
100 0.423
101 0.4165
102 0.4182
103 0.4157
104 0.4221
105 0.4205
106 0.4228
107 0.4243
108 0.4217
109 0.4236
110 0.4192
111 0.4201
112 0.4166
113 0.4231
114 0.4183
115 0.4185
116 0.4204
117 0.4234
118 0.4216
119 0.4222
120 0.4183
121 0.4245
122 0.4239
123 0.4248
124 0.4242
125 0.4233
126 0.4279
127 0.4171
128 0.4261
129 0.4269
130 0.4283
131 0.4239
132 0.4284
133 0.4279
134 0.4296
135 0.422
136 0.424
137 0.4264
138 0.4269
139 0.4263
140 0.4308
141 0.4267
142 0.4309
143 0.4265
144 0.4246
145 0.4227
146 0.4243
147 0.4165
148 0.4203
149 0.4262
150 0.4226
151 0.4308
152 0.4275
153 0.4238
154 0.4251
155 0.4285
156 0.4196
157 0.4308
158 0.4298
159 0.4295
160 0.4301
161 0.4267
162 0.4327
163 0.4261
164 0.4302
165 0.4295
166 0.4328
167 0.4324
168 0.4274
169 0.4289
170 0.4303
171 0.4294
172 0.4302
173 0.4271
174 0.4246
175 0.4248
176 0.432
177 0.4288
178 0.432
179 0.4323
180 0.4267
181 0.4222
182 0.4304
183 0.4293
184 0.4306
185 0.428
186 0.4315
187 0.43
188 0.4315
189 0.4234
190 0.4326
191 0.4294
192 0.4289
193 0.4271
194 0.43
195 0.4265
196 0.4233
197 0.4338
198 0.4279
199 0.4286
};
% \addlegendentry{WSGAN}
\end{axis}

\end{tikzpicture}
    % This file was created with tikzplotlib v0.10.1.
\begin{tikzpicture}

\definecolor{darkgray176}{RGB}{176,176,176}
\definecolor{darkorange25512714}{RGB}{255,127,14}
\definecolor{forestgreen4416044}{RGB}{44,160,44}
\definecolor{lightgray204}{RGB}{204,204,204}
\definecolor{steelblue31119180}{RGB}{31,119,180}

\begin{axis}[
legend cell align={left},
legend style={fill opacity=0.8, draw opacity=1, text opacity=1, draw=lightgray204, at={(1,0)}, anchor=south east},
tick align=outside,
tick pos=left,
title=\Huge\textbf{FashionMNIST},
x grid style={darkgray176},
xlabel=\Huge{Epoch},
xmajorgrids,
xmin=-9.95, xmax=208.95,
xtick style={color=black},
y grid style={darkgray176},
% ylabel=\Huge{ARI},
ymajorgrids,
ymin=-0.01611, ymax=0.8,
ytick style={color=black},
xlabel style={yshift=-14pt},
ylabel style={yshift=14pt},
font=\Huge
]
\addplot [line width=3, steelblue31119180]
table {%
0 0.0125
1 0.423
2 0.4903
3 0.5103
4 0.5232
5 0.5281
6 0.5323
7 0.5374
8 0.5371
9 0.5415
10 0.5405
11 0.5427
12 0.5461
13 0.5513
14 0.552
15 0.5454
16 0.552
17 0.5549
18 0.5504
19 0.5542
20 0.5571
21 0.557
22 0.5561
23 0.5593
24 0.5602
25 0.5575
26 0.56
27 0.5609
28 0.5661
29 0.5627
30 0.5645
31 0.5648
32 0.5663
33 0.5685
34 0.5684
35 0.5688
36 0.5653
37 0.5669
38 0.5692
39 0.569
40 0.5684
41 0.5681
42 0.5685
43 0.574
44 0.5689
45 0.5684
46 0.5667
47 0.5685
48 0.5688
49 0.5718
50 0.5717
51 0.5686
52 0.5712
53 0.5699
54 0.5679
55 0.5759
56 0.569
57 0.571
58 0.5735
59 0.5743
60 0.576
61 0.5731
62 0.5755
63 0.5782
64 0.5728
65 0.5764
66 0.5738
67 0.5773
68 0.5764
69 0.574
70 0.5728
71 0.5721
72 0.5703
73 0.5756
74 0.5711
75 0.5731
76 0.5726
77 0.5719
78 0.578
79 0.5747
80 0.5797
81 0.5771
82 0.5776
83 0.5789
84 0.5786
85 0.5718
86 0.5781
87 0.5778
88 0.5827
89 0.5786
90 0.5836
91 0.575
92 0.5801
93 0.571
94 0.5756
95 0.5769
96 0.581
97 0.5767
98 0.5789
99 0.5746
100 0.5766
101 0.576
102 0.5761
103 0.5781
104 0.5791
105 0.5784
106 0.5791
107 0.5764
108 0.5761
109 0.5778
110 0.581
111 0.5772
112 0.5804
113 0.5769
114 0.5798
115 0.5788
116 0.5796
117 0.5771
118 0.5763
119 0.5774
120 0.5806
121 0.5782
122 0.5779
123 0.5794
124 0.5776
125 0.5802
126 0.5766
127 0.5799
128 0.5756
129 0.5789
130 0.5786
131 0.5773
132 0.5818
133 0.5832
134 0.5786
135 0.5799
136 0.5829
137 0.5811
138 0.5775
139 0.5821
140 0.5799
141 0.5813
142 0.5818
143 0.5751
144 0.5782
145 0.5843
146 0.5764
147 0.5754
148 0.5762
149 0.5814
150 0.5804
151 0.5787
152 0.5786
153 0.5793
154 0.5773
155 0.5783
156 0.58
157 0.5813
158 0.5802
159 0.576
160 0.5821
161 0.5821
162 0.5799
163 0.5757
164 0.579
165 0.5814
166 0.5804
167 0.5809
168 0.5841
169 0.5804
170 0.5776
171 0.578
172 0.5768
173 0.5793
174 0.5815
175 0.5758
176 0.579
177 0.5793
178 0.5755
179 0.5792
180 0.5775
181 0.5774
182 0.5786
183 0.5783
184 0.5744
185 0.5778
186 0.5818
187 0.5796
188 0.5841
189 0.5788
190 0.5741
191 0.5762
192 0.5771
193 0.5801
194 0.5825
195 0.58
196 0.5816
197 0.5776
198 0.5795
199 0.5793
};
\addlegendentry{Ours(comp)}
\addplot [line width=3, darkorange25512714]
table {%
0 0.0143
1 0.4198
2 0.4842
3 0.5044
4 0.5132
5 0.5271
6 0.5332
7 0.5363
8 0.5363
9 0.5376
10 0.5392
11 0.5458
12 0.5437
13 0.5466
14 0.5483
15 0.5541
16 0.5529
17 0.5527
18 0.55
19 0.5581
20 0.5601
21 0.5531
22 0.5564
23 0.5535
24 0.5555
25 0.5589
26 0.5555
27 0.5618
28 0.5588
29 0.5644
30 0.564
31 0.5598
32 0.5602
33 0.5663
34 0.5627
35 0.5668
36 0.5645
37 0.5672
38 0.5696
39 0.5674
40 0.5719
41 0.5706
42 0.5659
43 0.5675
44 0.5712
45 0.5687
46 0.5659
47 0.5659
48 0.5676
49 0.5688
50 0.5689
51 0.5703
52 0.5751
53 0.572
54 0.5706
55 0.5732
56 0.572
57 0.5731
58 0.5716
59 0.5698
60 0.5717
61 0.5725
62 0.5674
63 0.5744
64 0.5713
65 0.568
66 0.5728
67 0.5711
68 0.5736
69 0.5736
70 0.575
71 0.5761
72 0.5715
73 0.5774
74 0.5727
75 0.5733
76 0.572
77 0.5729
78 0.5707
79 0.5727
80 0.5754
81 0.5712
82 0.5764
83 0.5748
84 0.5775
85 0.5717
86 0.575
87 0.5779
88 0.5736
89 0.5716
90 0.5771
91 0.5772
92 0.5746
93 0.5763
94 0.5821
95 0.5776
96 0.577
97 0.5776
98 0.5793
99 0.5808
100 0.5755
101 0.5779
102 0.5768
103 0.5771
104 0.578
105 0.58
106 0.579
107 0.5763
108 0.5765
109 0.5758
110 0.5771
111 0.5781
112 0.579
113 0.5778
114 0.5731
115 0.5742
116 0.5766
117 0.5774
118 0.5784
119 0.573
120 0.5756
121 0.5802
122 0.5776
123 0.5798
124 0.5754
125 0.576
126 0.5766
127 0.5754
128 0.5795
129 0.5765
130 0.5751
131 0.579
132 0.5798
133 0.5729
134 0.5738
135 0.573
136 0.5785
137 0.5823
138 0.5833
139 0.5793
140 0.5785
141 0.5744
142 0.5816
143 0.577
144 0.5786
145 0.5733
146 0.5802
147 0.5812
148 0.579
149 0.5844
150 0.5794
151 0.5748
152 0.5785
153 0.5753
154 0.5815
155 0.5815
156 0.5801
157 0.5784
158 0.5812
159 0.5801
160 0.5825
161 0.5803
162 0.584
163 0.578
164 0.5776
165 0.58
166 0.5748
167 0.5782
168 0.5775
169 0.5826
170 0.5824
171 0.5847
172 0.5806
173 0.5802
174 0.5784
175 0.5779
176 0.5826
177 0.5798
178 0.5812
179 0.5787
180 0.5812
181 0.582
182 0.5755
183 0.5776
184 0.5793
185 0.5816
186 0.5811
187 0.5797
188 0.5775
189 0.5821
190 0.5809
191 0.5847
192 0.5805
193 0.5783
194 0.5797
195 0.5791
196 0.5758
197 0.5787
198 0.5803
199 0.5826
};
\addlegendentry{Ours(sub)}
\addplot [line width=3, forestgreen4416044]
table {%
0 0.2764
1 0.385
2 0.4547
3 0.4927
4 0.5069
5 0.5112
6 0.5216
7 0.5254
8 0.5236
9 0.5267
10 0.532
11 0.5334
12 0.5309
13 0.5277
14 0.5262
15 0.5241
16 0.5179
17 0.5284
18 0.5248
19 0.5177
20 0.5205
21 0.5241
22 0.5195
23 0.5249
24 0.5278
25 0.5263
26 0.5335
27 0.5313
28 0.5304
29 0.5343
30 0.5385
31 0.5359
32 0.5333
33 0.5264
34 0.5328
35 0.5333
36 0.5393
37 0.5342
38 0.5322
39 0.5348
40 0.5404
41 0.5363
42 0.5394
43 0.5378
44 0.5324
45 0.5366
46 0.534
47 0.5392
48 0.5395
49 0.5371
50 0.5431
51 0.5389
52 0.5361
53 0.5327
54 0.5336
55 0.5302
56 0.5359
57 0.5349
58 0.5346
59 0.5365
60 0.5365
61 0.531
62 0.5368
63 0.5332
64 0.5403
65 0.5305
66 0.5319
67 0.5321
68 0.5326
69 0.5318
70 0.5301
71 0.5311
72 0.534
73 0.5327
74 0.5298
75 0.533
76 0.5315
77 0.5315
78 0.5273
79 0.5329
80 0.5272
81 0.5254
82 0.5288
83 0.5291
84 0.5302
85 0.5278
86 0.5239
87 0.5254
88 0.5259
89 0.5286
90 0.5247
91 0.5255
92 0.5204
93 0.522
94 0.5206
95 0.5208
96 0.5249
97 0.5188
98 0.5209
99 0.5229
100 0.5193
101 0.5229
102 0.5201
103 0.52
104 0.519
105 0.5203
106 0.5274
107 0.517
108 0.5181
109 0.5188
110 0.5131
111 0.5179
112 0.5212
113 0.5209
114 0.5178
115 0.5185
116 0.5193
117 0.5163
118 0.5178
119 0.5136
120 0.5143
121 0.519
122 0.5221
123 0.518
124 0.5127
125 0.5267
126 0.5141
127 0.5159
128 0.5174
129 0.5156
130 0.5165
131 0.5156
132 0.509
133 0.5143
134 0.5135
135 0.512
136 0.5148
137 0.5181
138 0.5201
139 0.5179
140 0.5119
141 0.5174
142 0.5175
143 0.5145
144 0.5099
145 0.5166
146 0.5162
147 0.5153
148 0.5216
149 0.5242
150 0.5205
151 0.5157
152 0.5119
153 0.5155
154 0.5125
155 0.5213
156 0.5114
157 0.5169
158 0.5148
159 0.517
160 0.5156
161 0.5133
162 0.5162
163 0.5125
164 0.5184
165 0.5166
166 0.5133
167 0.5124
168 0.517
169 0.5104
170 0.5116
171 0.5131
172 0.5107
173 0.5118
174 0.5174
175 0.5172
176 0.5128
177 0.5105
178 0.5064
179 0.5105
180 0.5138
181 0.5144
182 0.5054
183 0.5089
184 0.5058
185 0.5079
186 0.5056
187 0.5034
188 0.5058
189 0.5006
190 0.5075
191 0.5055
192 0.511
193 0.5059
194 0.5129
195 0.4979
196 0.5055
197 0.5017
198 0.5041
199 0.4995
};
\addlegendentry{WSGAN}
\end{axis}

\end{tikzpicture}
}
\caption{Comparison of Adjusted Rank Index (ARI) of proposed methods and WSGAN.}
    \label{fig:ARI}
\end{figure}
% \begin{figure}[ht!]
%     \centering
%      \resizebox{0.15\textwidth}{!}
%  {
%   % \input{AAAI/ari_plots/cifar10_a+v-5_b--1.0_a-None+v-3_b-0.9_a-2.0+v-3_b-None_a-None}    
%   %       \input{AAAI/ari_plots/cifar10_b+v-3_b--1.0_a-None+v-2_b-0.9_a-2.5+v-2_b-None_a-None}
%          \input{AAAI/ari_plots/fashion_mnist+v-3_b--1.0_a-None+v-2_b-0.9_a-2.0+v-4_b-None_a-None}
% }
% \caption{Comparison of Adjusted Rank index(ARI) of proposed methods and WSGAN.}
%     \label{fig:ARI}
% \end{figure}

% For this, we leverage the images generated by the GAN and their corresponding pseudo labels in addition to the training dataset for the training of ResNet-18 model and report the percentage increase in the test accuracy.
To show the improvement in the accuracy of the classifier, we have reported the Adjusted Rank Index score (ARI) for the training epochs. We have conducted a comparative analysis with different label models, including Majority Voting (MV), MeTaL~\cite{PWS_ratner2019_metal}, FlyingSquid (FS)~\cite{PWS_fu_fastsquid}, Snorkel~\cite{PWS_ratner2020snorkel}, hyper-label model(HLM)~\cite{wu2022learning} and DawidSkene (DS)~\cite{PWS_dawid1979}. To facilitate this comparison, we employed the label model codebase from Wrench~\cite{wrench} and utilized the official codebase provided by WSGAN and hyper-label model. If any conflicts emerged during the implementation process due to version disparities, we referred to the numerical values reported in the paper. When information was unavailable, we omitted to report those particular numbers. 
% \vspace{-2mm} %% cutting down extra space
\subsection{Results on the classifier performance}
% To demonstrate the effectiveness of the classifier in discovering the underlying class label. 
Figure-\ref{fig:ARI} shows the Adjusted Rank Index between the prediction made by our classifier and the underlying ground truth label. We have further compared it with the prediction made by WSGAN. The results indicate that the noise-aware classifier is more effective than WSGAN in learning the underlying ground truth labels.
\subsection{Results on label model performance}
Table-\ref{tab:label_model_acc} shows the comparison of the mean posterior accuracy of our proposed method with other baselines. Compared to WSGAN, our proposed method gives a 3.0\% improvement in AWA2 dataset and around 1.8\% improvement in CIFAR10-A. Among the subset selection and the non-abstained data, we found a marginal difference in the performance of the label model.

\subsection{Results on image generation}
The results in Table-\ref{Tab:FID} highlight the average FID scores achieved by our methods compared to those of WSGAN. On average, our method achieves 7.08 point FID improvement compared to WSGAN. Additionally, we observe that employing a subset selection scheme for the classifier further improves the quality of generated images. This distinction is particularly notable for the GTSRB dataset, with a 3.46-point FID enhancement.

\subsection{Results on data augmentation}
To examine the improvement in test accuracy for the data augmentation task, we trained a ResNet-18~\cite{resnet} model using around 1000 synthetic images generated by the GAN, in addition to the original training dataset.
 For pseudo labels, we have considered two settings. Firstly, we use the conditional discrete latent vector (L) used for image generation. Secondly, we use the output of the classifier to generate the associated class label for the images (CL). The results of the experiment is reported in Table-\ref{tab:resnet}. Comparatively, we found that our proposed method generates better results than the baseline. Among the subset selection scheme and the non-abstained data, we found the performance of the subset selection scheme to be better for DomainNet, FashionMNIST, CIFAR10-B, CIFAR10-A for conditional discrete latent vector (L), and AWA2, CIFAR10-A, GTSRB, and CIFAR10-B for classifier based pseudo labels (CL).     

% \subsection{Performance Comparison}
%     In order to cement our claim that through our class conditioned based approach to image generation results in better images than previous approaches, we train a simple ResNet-18\cite{resnet} classifier on each of the datasets previously mentioned, by considering at most 1000 generated images for that particular dataset. Train and test split for AWA2 and DomainNet datasets was [0.85, 0.15], for all other datasets the PyTorch split was used. The test data was further split equally into test and validation sets. The test set was kept constant by setting the seed and the initialization of weights for all models was also seeded to ensure like-for-like comparison. For each dataset, we train four classifier models on:
%     \begin{enumerate}
%         \item Original dataset
%         \item Original dataset + images generated by WSGAN
%         \item Original dataset + images generated by our approach(complete data)
%         \item Original dataset + images generated by our approach(submodularity)
%     \end{enumerate}
%     The percentage increase in accuracy on the test set is reported in Table \ref{tab:resnet}.
% \subsection{LabelModels}
%      Our results are in close alignment with the findings of \cite{gen_model_PWS_vice_versa}, but we weren't able to Table \ref{tab:label_model_acc} shows the performance of the models on various datasets with posterior accuracy as the metric, where we find that our approach consistently outperforms other approaches.

\section{Conclusions and Acknowledgments}
In this work, we introduced a new framework to fuse the training of conditional GAN and label model using a noise-aware weakly supervised classifier. We have further investigated the impact of training the classifier using a subset of representative and diverse samples with limited uncertainty associated with the pseudo labels. Our empirical results indicate an improvement in the quality of images and the accuracy of pseudo labels compared to existing state-of-the-art techniques. We further demonstrate that employing a subset selection strategy helps in improving the image quality of GAN, and a small set of training data can be used to generate comparable performance of label model.  It is worth noting that the subset selection approach has limitations tied to the size of the non-abstained sample set, which can result in computational challenges for very large datasets. In future work, we aim to extend our approach to bigger datasets. \\
This work was supported (in part for setting up the GPU compute)
by the Indian Institute of Science through a start-up grant. Prathosh
is supported by the Infosys Foundation Young Investigator Award.

\bibliography{aaai24}
\onecolumn
\section{Supplemantary Material}
\subsection{Architecture}
For performance comparison, we used the DCGAN architecture~\cite{DCGAN} for 32X32 images. Our classifier network shares all the convolutional layers with the discriminator and uses a separate, fully connected layer to predict the associated class for a given image. Further, we have considered a 100-dimensional latent vector for image generation and sample the class conditional vector from a uniform distribution.

For the accuracy parameters associated with the label model($\gA$), we have used the image features extracted from our shared convolutional network. The accuracy parameters are modeled using an MLP network with the same output dimension as the number of label functions used for the experiment. The network consists of three layers of feed-forward network with dimensions [256, 128, 64] and uses a ReLU activation for intermediate layers and Sigmoid for the final prediction layer. The fully connected layer $\gF$ used in training the classifier consists of a simple linear layer with softmax output. 

% \begin{figure}[h]
%     \centering
%     \includegraphics{AAAI/block_diagram.svg}
%     \caption{Architecture}
%     \label{fig:architecture}
% \end{figure}

\subsection{Training}

The complete model is trained for 200 epochs using five Adam optimizers, one for each loss function that comprises adversarial loss for the generator,  adversarial loss for the discriminator, loss associated with the label model, classifier loss, and log-likelihood maximization-based loss for the generator. The learning rate associated with each of these optimizers are $1\text{x}10^{-4}$, $4\text{x}10^{-4}$ , $8\text{x}10^{-5}$, $1.8\text{x}10^{-4}$, $1\text{x}10^{-5}$ respectively. We have used a weight of 0.7 for cross entropy and 0.3 for reverse cross-entropy loss for the soft cross-entropy loss.
In our subset selection-based experiment, we normalized the entropy to ensure that the total entropy $(Cost(\Tilde{D}_t))$ equals the cardinality of the non-abstained samples. This normalization was necessary to facilitate the use of the apricot library for knapsack-based subset selection. 
We performed a grid search for other hyper-parameters associated with subset selection to identify the best for all the datasets. Specifically, for entropy budget ($\eta$), we considered a range of values such as [0.7, 0.8, 0.9], and for the trade-off between representativeness and diversity in generalized graph cut ($\gamma$), we explored values such as [2, 2.5, 3, 3.5]. In $\vl_{decay}$ loss for the hyper-parameter $\mu$ we have formulated it as $\frac{\gC}{e*\delta+1}$~\cite{gen_model_PWS_vice_versa}, where $\gC$ is the total number class in the data, $e$ is the current epoch of training and $\delta$ is set to be 1 for experiments across all datasets.  Further, the label model and classifier were trained using an appropriate stop gradient operator on the classifier's prediction and pseudo labels, respectively. The pseudo labels for the training of the classifier are updated after every epoch.

\section{Algorithm}
In Algorithm -~\ref{alg:proposed}, we describe our proposed method for training the label model and the conditional GAN. 
\begin{algorithm}

\caption{Pseudo code of our proposed method. }
\label{alg:proposed}
   \begin{algorithmic}
   \State\textbf{Input} training dataset $(\Tilde{\gD})$, Label functions ($\Lambda$), label model ($\gL$), non-abstained dataset $(\Tilde{\gD}_t)$, hyper-parameter($p, \eta, \omega, \gB$) 
    \For{ i in [0...\textit{epoch}] }
        \If{i\%p == 0} 
    \State  \textcolor{gray}{\textbf{\textit{See (Eq. 10) for CEG Formulation.}}}
    \State $\gS_{cost}$ = CEG($\Tilde{\gD_t},\gB, c= entropy \text { } function (.)$) \textit{\textcolor{gray}{\# final set selected using entropy-based cost function (Eq. 9)}}. 
    \State $\gS_{uniform}$ = CEG($\Tilde{D_t},\gB, c = 1$) \textit{\textcolor{gray}{\# final set selected using uniform cost}}. 
    \State $\Tilde{\gD}_0 = \argmax_{\gS} (\gF(\gS_{uniform}),\gF(\gS_{Cost} ))))$ \textit{\textcolor{gray}{\# select the set with maximum utility (Eq. 7). }}
    \State $\hat{y} = 
\argmax  \gL(\Lambda(x_i);\theta_{\gA}^i)$ where $x_i \in \Tilde{\gD}_0$, $\hat{\gY} =\{\hat{y_1} \hdots \hat{y}_{\Tilde{|\gD_0|}} \}$ \textit{\textcolor{gray}{\# pseudo labels generated by label model  (Eq. 4).}}
    \EndIf
    \For {b in training batches : }
    \State $\theta_{\gG}^{i+1} = \theta_{\gG}^{i} -  \omega_0 \nabla(\vl_{gan}(\gG, \gD, \gC ; \theta_{\gG}^i,\theta_{\gD}^i, \theta_{\gC}^i))$  \textit{\textcolor{gray}{\# training the parameters of generator (Eq. 2).}}
    \State $\theta_{\gD}^{i+1} = \theta_{\gD}^{i} +  \omega_1 \nabla(\vl_{gan}(\gG, \gD, \gC ; \theta_{\gG}^i,\theta_{\gD}^i, \theta_{\gC}^i))$ \textit\textit{\textcolor{gray}{\# training the parameters of discriminator (Eq. 2).}}
    \State $\theta_{\gA}^{i+1} = \theta_{\gA}^{i} -  \omega_2 \nabla(\vl_{lm}(\gA; \theta_{\gA}^i))$ \textit{\textcolor{gray}{\# training the parameters of label model (Eq. 5, 6).}}
    \State $\theta_{\gC}^{i+1} = \theta_{\gC}^{i} -  \omega_3 \nabla(\vl_{sce}(\Tilde{\gD}_0,\hat{\gY}; \theta_\gC^i))$ \textit{\textcolor{gray}{\# training the parameters of classifier(Eq. 1).}}
    \EndFor
    \EndFor
    \end{algorithmic}
\end{algorithm}
\subsection{Subset selection of data under the knapsack constraint}
\begin{prop} \cite{leskovec2007cost,submodular_knapsack_NAS}
     If $\mathcal{\gF}$ is a non decreasing submodular function with $\gF\big(\phi\big)$ = 0 then CEG algorithm achieves a constant ratio of  $\frac{1}{2} \big(1- \frac{1}{e}\big)$  to the optimal solution under the knapsack constraints.   
     \begin{align}
     \max \big(\gF({\gS}_{uniform}),
     \gF({\gS}_{cost})\big)\geq \frac{1}{2} \big(1- \frac{1}{e}\big)\max_{Cost(\gS)\leq \gB } \gF(\gS) \nonumber
     \end{align}
     \label{prop:submod}
 \end{prop}
 In our setting, $\gF({\gS}_{uniform}$) and $\gF({\gS}_{cost}$) are the final subsets selected using the cost-effective greedy algorithm, and $\gF$ is the generalized graph cut algorithm. 
% \clearpage
\section{Network Ablation using StyleGAN-ADA}
To show the compatibility of our proposed method with different architectures, we trained styleGAN-ADA~\cite{styleGAN-ADA} on CIFAR10 images using self-supervised based label functions (CIFAR10-B). The implementation uses the official StyleGAN-ADA codebase with a setup similar to previous work~\cite{gen_model_PWS_vice_versa}. However, we have used a depth mapping of 7 and an embedding size of 150 for all the experiments related to CIFAR10.

\begin{figure}[ht!]
        \centering

        \includegraphics[width=350px]{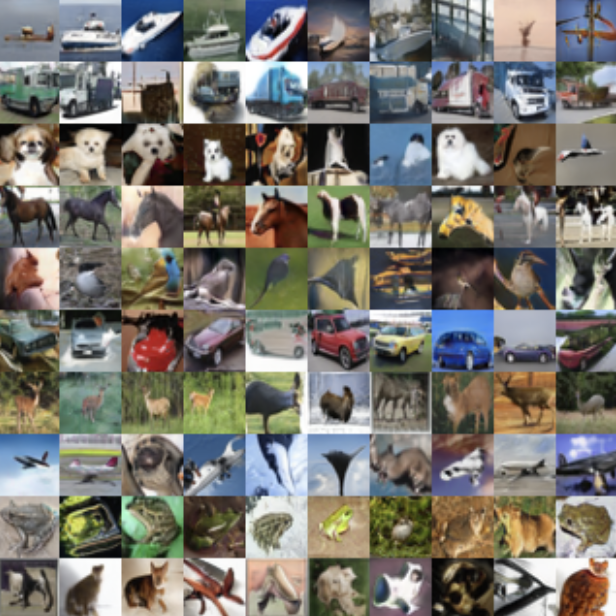}
        \caption{Images generated by our method using StyleGAN-ADA based architecture on CIFAR10-B.}
        \label{fig:cifar_stylegan}
    \end{figure}

Further, we performed these experiments on a single GPU because of limited computing resources. For CIFAR10-B, we run the model until the discriminator has seen around 19M real images. For the setting where the complete set of non-abstained samples ($\Tilde{\gD_0} = \Tilde{\gD_t}$) was used for the training of the classifier, we achieved an FID of 4.96 and an accuracy of 73.31\% for label model. On the other hand, using the subset selection method, we achieved an FID of 4.94 and an accuracy of 73.18\%.  Figure -~\ref{fig:cifar_stylegan} shows the images generated by our model.  

    \begin{figure}
        \centering
        \includegraphics[width=0.5\linewidth]{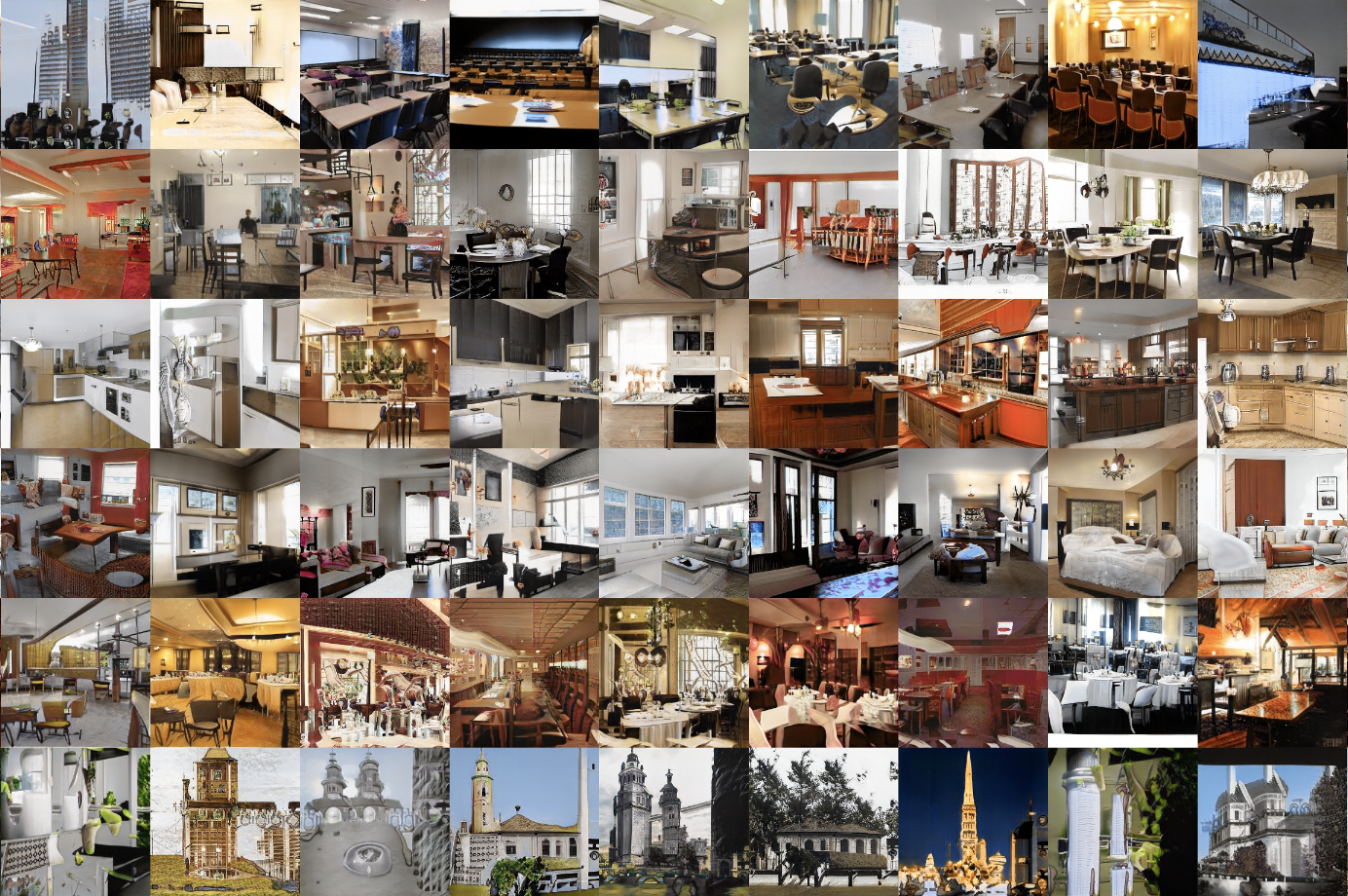}
        \caption{Images generated by our method using StyleGAN-ADA based architecture on LSUN.}
        \label{fig:lsun_stylegan}
    \end{figure}

\section{Results on High-Resolution Images}
To demonstrate the effectiveness of our method on high-resolution images, we experimented with LSUN~\cite{yu2015lsun} (256x256, 10 class, self-supervised learning based LFs) and AFHQ~\cite{choi2020stargan} (256x256, 3 class, synthetic LFs) datasets containing 100000 and 14629 samples, respectively. The LFs for LSUN are randomly sampled from the original ssl-based LFs provided by \citet{gen_model_PWS_vice_versa}. The styleGAN-ADA-based~\cite{styleGAN-ADA} network was employed for both of these datasets. Results in Table-\ref{tab:rebuttal_fid} demonstrate the efficacy of our method on generation quality and label model performance on a high-resolution dataset.
\begin{table*}[ht!]
\centering
\resizebox{0.73\linewidth}{!}
{
    \begin{tabular}{ l | c | c | c | c | c | c }
    \hline
     & \multicolumn{2}{c |}{\textbf{WSGAN}} & \multicolumn{2}{c |}{\textbf{Ours(comp)}} & \multicolumn{2}{c }{\textbf{Ours(sub)}} \\
    \hline
    \textbf{Dataset} & \textbf{FID} & \textbf{Accuracy} & \textbf{FID} & \textbf{Accuracy} & \textbf{FID} & \textbf{Accuracy} \\
    \hline
    LSUN & 14.47 & 0.765 & 17.42 & 0.773 & \textbf{14.24} & \textbf{0.777} \\
    
    % LSUN-5 &  &  &  &  &  &  \\
    AFHQ & 24.72 & \textbf{0.823}  & 24.31 & 0.815  & \textbf{23.13} & 0.815 \\
    \hline
    \end{tabular}
}
\caption{FID \& train posterior accuracy for high-resolution images}
\label{tab:rebuttal_fid}
\end{table*}

\section{Sensitivity of subset selection based hyper-parameters on the performance of the model}

    To better understand the impact of different hyper-parameters on the performance of the label model and the quality of images generated by the GAN. We conducted a sensitivity test on the hyper-parameters, such as the budget ratio on knapsack constraint($\eta$) and the tradeoff between representativeness and diversity($\gamma$) of the selected samples in generalized graph cut. For these experiments, we have used optimal values for the remaining hyper-parameters.    
    \subsection{Sensitivity test on knapsack constraint ($\eta$)}
        In the given experiment, we fixed the value of $\gamma$ equal to 3.0 and varied the hyper-parameter associated with the knapsack constraint ($\eta$) between a range of 0.2 to 0.8 with 0.2 as the step size. Table - \ref{tab:eta_FID} shows the result of the given experiment on the quality of images generated by the conditional GAN. Similarly, Table - \ref{tab:eta_acc} shows the performance of the label model on a non-abstained dataset. 
        
        For the FID score, we found that for datasets like GTSRB, MNIST, and CIFAR10-B, the best performance was generated for a smaller subset of data. However, for AWA2, CIFAR10-A, and FashionMNIST, a bigger data set ($\eta=0.8$) generates the best results.  
        For posterior accuracy, we found that for datasets like AWA2, CIFAR10-A, CIFAR10-B, and FashionMNIST, the model's performance increases as the subset size increases. However, the performance of a smaller subset of data is still comparable to the best results generated during the experiment.

        \begin{table*}[!ht]
            \centering
            \resizebox{\textwidth}{!}{
                \begin{tabular}{l|lllllll}
                \hline
                    \textbf{$\eta$} & \textbf{AWA2} & \textbf{CIFAR10-A} & \textbf{CIFAR10-B} & \textbf{DomainNet} & \textbf{FashionMNIST} & \textbf{GTSRB} & \textbf{MNIST} \\ \hline
                    0.2 & 29.207 & 20.696 & 19.972 & 40.690 & 13.686 & 61.260 & 4.202 \\
                    0.4 & 29.683 & 20.175 & 19.702 & 40.861 & 13.080 & \textbf{56.443} & \textbf{2.732} \\
                    0.6 & 29.744 & 19.606 & \textbf{19.000} & 41.250 & 12.542 & 60.255 & 4.495 \\
                    0.8 & \textbf{28.956} & \textbf{18.331} & 19.578 & \textbf{37.709} & \textbf{11.747} & 59.590 & 4.838 \\
                    \hline
                \end{tabular}
            }
            \caption{Sensitivity test on the hyper-parameter $\eta$ over FID. This experiment illustrates the impact of selecting subsets of different sizes on the quality of images generated by GAN. The best score for a dataset is highlighted in \textbf{bold}.}
            \label{tab:eta_FID}
        \end{table*}
        
        \begin{table*}[!ht]
            \centering
            \resizebox{\textwidth}{!}{
                \begin{tabular}{l|lllllll}
                \hline 
                    \textbf{$\eta$} & \textbf{AWA2} & \textbf{CIFAR10-A} & \textbf{CIFAR10-B} & \textbf{DomainNet} & \textbf{FashionMNIST} & \textbf{GTSRB} & \textbf{MNIST} \\ \hline
                    0.2 & 0.672 & 0.873 & 0.731 & 0.633 & 0.739 & 0.824 & 0.815 \\
                    0.4 & 0.684 & 0.883 & 0.735 & 0.640 & 0.750 & 0.827 & 0.820 \\
                    0.6 & 0.693 & 0.887 & 0.738 & \textbf{0.647} & 0.752 & \textbf{0.829} & \textbf{0.822} \\
                    0.8 & \textbf{0.698} & \textbf{0.889} & \textbf{0.739} & 0.644 & \textbf{0.753} & 0.828 & 0.820 \\ \hline
                \end{tabular}
            }
            \caption{Sensitivity test on the hyper-parameter $\eta$ over average posterior accuracy. This experiment illustrates the impact of selecting subsets of different sizes on the average posterior accuracy of the label models for samples with at least one vote from the label function. The best score for a specific dataset is highlighted in \textbf{bold}.}
            \label{tab:eta_acc}
        \end{table*}

      \begin{table*}[!h]
            \centering
            \resizebox{\textwidth}{!}{
                \begin{tabular}{l|lllllll}
                \hline 
                    \textbf{$\gamma$} & \textbf{AWA2} & \textbf{CIFAR10-A} & \textbf{CIFAR10-B} & \textbf{DomainNet} & \textbf{FashionMNIST} & \textbf{GTSRB} & \textbf{MNIST} \\ \hline
                    2.0 & 25.439 & 20.220 & 18.499 & 37.368 & \textbf{10.533} & \textbf{55.710} & 4.323 \\
                    2.5 & 26.061 & 19.913 & \textbf{18.159} & 36.561 & 11.293 & 61.622 & 6.013 \\
                    3.0 & 26.016 & \textbf{17.502} & 18.938 & \textbf{36.008} & 11.293 & 56.494 & \textbf{3.494} \\
                    3.5 & \textbf{23.163} & 18.933 & 18.478 & 36.456 & 11.971 & 57.574 & 6.264 \\ \hline
                \end{tabular}
            }
            \caption{Sensitivity test on the hyper-parameter $\gamma$ over FID. This experiment illustrates the impact of the trade-off between representativeness and diversity of the selected subset on the quality of images generated by GAN. The best score for a dataset is highlighted in \textbf{bold}.}
            \label{tab:gamma_FID}
        \end{table*}
        
        \begin{table*}[!ht]
            \centering
            \resizebox{\textwidth}{!}{
                \begin{tabular}{l|lllllll}
                \hline
                    \textbf{$\gamma$} & \textbf{AWA2} & \textbf{CIFAR10-A} & \textbf{CIFAR10-B} & \textbf{DomainNet} & \textbf{FashionMNIST} & \textbf{GTSRB} & \textbf{MNIST} \\ \hline
                    2.0 & 0.702 & 0.888 & 0.740 & 0.652 & \textbf{0.754} & 0.828 & \textbf{0.821} \\
                    2.5 & 0.702 & 0.888 & 0.740 & \textbf{0.655} & \textbf{0.754} & \textbf{0.829} & \textbf{0.821} \\
                    3.0 & \textbf{0.703} & 0.888 & 0.740 & 0.651 & \textbf{0.754} & 0.828 & \textbf{0.821} \\
                    3.5 & \textbf{0.703} & \textbf{0.889} & \textbf{0.741} & 0.650 & 0.753 & \textbf{0.829} & 0.820 \\ \hline
                \end{tabular}
            }
            \caption{Sensitivity test on the hyper-parameter $\gamma$ over average posterior accuracy. This experiment illustrates the impact of selecting representative and diverse subsets on the average posterior accuracy of the label models for samples with at least one vote from the label function. The best score for a dataset is highlighted in \textbf{bold}.}
            \label{tab:gamma_acc}
        \end{table*}
        
    \subsection{Sensitivity test on generalized graph cut($\gamma$) }
        
        We have further conducted a sensitivity test on the hyper-parameter $\gamma$ to understand the impact of the tradeoff on representativeness and diversity for selected samples on the performance of the model. For this experiment, the entropy budget ratio ($\eta$) was fixed at 0.8, and the hyperparameter $\gamma$ was varied between 2 and 3.5 at a step size of 0.5. Table-\ref{tab:gamma_FID} and Table-\ref{tab:gamma_acc} illustrate the impact of the given hyper-parameter on the performance of conditional GAN and label model. 

        We can clearly see the impact of the given hyperparameter on the FID score. However, the label model's performance remains relatively stable with a change in these values.
      
% \clearpage
\section{Comparison of baseline on different metrics}
Table - \ref{tab:F1}, Table-\ref{tab:recall} and Table-\ref{tab:precision} show the F1-score, Recall, and Precision achieved by our methods in comparison to other label models. The score clearly illustrates that our methods generate better performance than other baselines.    
\begin{table*}[!ht]
    \centering
    \resizebox{\textwidth}{!}{
        \begin{tabular}{l|llllllll}
            \hline
            \textbf{Dataset} & \textbf{MV} & \textbf{MeTaL} & \textbf{FS} & \textbf{Snorkel} & \textbf{DS} & \textbf{WSGAN} & \textbf{Ours(comp)} & \textbf{Ours(sub)} \\ \hline
            AWA2 & 0.609 & 0.569 & 0.596 & 0.581 & - & 0.652 & \textbf{0.683} & 0.678 \\
            CIFAR10-A & 0.828 & 0.798 & 0.795 & 0.798 & 0.850 & 0.870 & \textbf{0.887} & \textbf{0.887} \\
            CIFAR10-B & 0.715 & 0.703 & 0.703 & 0.704 & 0.672 & 0.725 & \textbf{0.736} & 0.735 \\
            DomainNet & 0.581 & 0.446 & 0.622 & 0.436 & \textbf{0.655} & 0.632 & 0.642 & 0.641 \\
            FashionMNIST & 0.704 & 0.698 & 0.705 & 0.703 & 0.691 & 0.715 & \textbf{0.724} & \textbf{0.724} \\
            GTSRB & 0.802 & 0.800 & 0.628 & 0.799 & 0.616 & 0.811 & \textbf{0.814} & \textbf{0.814} \\
            MNIST & 0.759 & 0.738 & 0.755 & 0.744 & 0.716 & 0.798 & \textbf{0.800} & \textbf{0.800} \\ \hline
        \end{tabular}
    }
    \caption{Comparison between the F1 scores of the label models for samples with at least one vote from the label function. The best performer across methods is highlighted in \textbf{bold}.}
    \label{tab:F1}
\end{table*}

\begin{table*}[!ht]
    \centering
    \resizebox{\textwidth}{!}{
        \begin{tabular}{l|llllllll}
            \hline
            \textbf{Dataset} & \textbf{MV} & \textbf{MeTaL} & \textbf{FS} & \textbf{Snorkel} & \textbf{DS} & \textbf{WSGAN} & \textbf{Ours(comp)} & \textbf{Ours(sub)} \\ \hline
            AWA2 & 0.627 & 0.619 & 0.621 & 0.626 & - & 0.670 & \textbf{0.701} & 0.696 \\
            CIFAR10-A & 0.831 & 0.804 & 0.800 & 0.804 & - & 0.872 & \textbf{0.890} & 0.889 \\
            CIFAR10-B & 0.719 & 0.708 & 0.708 & 0.709 & - & 0.729 & \textbf{0.740} & 0.739 \\
            DomainNet & 0.595 & 0.485 & 0.635 & 0.478 & - & 0.642 & \textbf{0.650} & 0.649 \\
            FashionMNIST & 0.735 & 0.730 & 0.734 & 0.734 & - & 0.744 & \textbf{0.754} & \textbf{0.754} \\
            GTSRB & - & - & - & - & - & 0.825 & \textbf{0.828} & \textbf{0.828} \\
            MNIST & 0.778 & 0.759 & 0.773 & 0.764 & - & 0.816 & \textbf{0.818} & \textbf{0.818} \\
        \hline
        \end{tabular}
    }
    \caption{Comparison between the Recall scores of the label models for samples with at least one vote from the label function. The best performer across methods is highlighted in \textbf{bold}.}
    \label{tab:recall}
\end{table*}

\begin{table*}[!ht]
    \centering
    \resizebox{\textwidth}{!}{
        \begin{tabular}{l|llllllll}
            \hline
            \textbf{Dataset} & \textbf{MV} & \textbf{MeTaL} & \textbf{FS} & \textbf{Snorkel} & \textbf{DS} & \textbf{WSGAN} & \textbf{Ours(comp)} & \textbf{Ours(sub)} \\ \hline
            AWA2 & 0.644 & 0.637 & 0.650 & 0.629 & - & 0.674 & \textbf{0.709} & 0.705 \\
            CIFAR10-A & 0.839 & 0.823 & 0.819 & 0.821 & 0.866 & 0.875 & \textbf{0.892} & \textbf{0.892} \\
            CIFAR10-B & 0.738 & 0.736 & 0.736 & 0.737 & 0.659 & 0.746 & \textbf{0.756} & 0.755 \\
            DomainNet & 0.667 & 0.700 & 0.676 & 0.697 & \textbf{0.702} & 0.682 & 0.694 & 0.692 \\
            FashionMNIST & 0.687 & 0.688 & 0.690 & 0.711 & \textbf{0.712} & 0.699 & 0.707 & 0.707 \\
            GTSRB & 0.718 & 0.761 & 0.714 & 0.772 & 0.731 & 0.809 & \textbf{0.812} & 0.811 \\
            MNIST & 0.753 & 0.744 & 0.749 & 0.742 & 0.772 & 0.787 & \textbf{0.789} & \textbf{0.789} \\
        \hline
        \end{tabular}
    }
    \caption{Comparison between the Precision scores of the label models for samples with at least one vote from the label function. The best performer across methods is highlighted in \textbf{bold}.}
    \label{tab:precision}
\end{table*}

\section{Generated Images}
\begin{figure}[ht!]
        \centering
        \captionsetup{width=.7\linewidth}
        \includegraphics[width=250px]{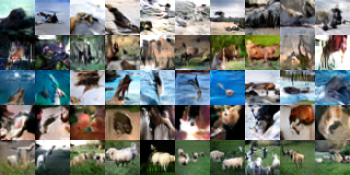}
        \caption{AWA2 images generated by DCGAN that utilizes our proposed method. It is to be noted that the generation of images is challenging due to the small size of our dataset (around 7000 images).}
        \label{fig:gen_awa2}
    \end{figure}

\begin{figure}[ht!]
        \centering
        \captionsetup{width=.7\linewidth}
        \includegraphics[width=250px]{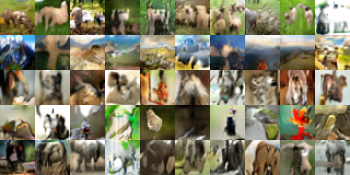}
        \caption{DomainNet images generated by DCGAN that utilizes our proposed method. It is to be noted that the generation of images is challenging due to the small size of our dataset (around 7000 images).}
        \label{fig:gen_domain_net}
    \end{figure}

 This section presents the images generated through our approach with a DCGAN architecture. For each dataset, five classes were randomly chosen, and for each of the classes, ten randomly chosen images were shown. For AWA2 (Figure-\ref{fig:gen_awa2}) and DomainNet(Figure-\ref{fig:gen_domain_net}) the complete training process uses around 7000 samples, which makes the generation of images relatively harder in comparison to other datasets.

    \begin{figure}[ht!]
        \centering
        \captionsetup{width=.7\linewidth}
        \includegraphics[width=250px]{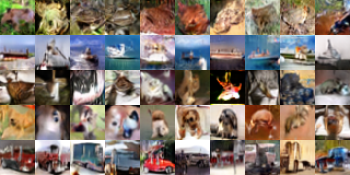}
        \caption{CIFAR10 images generated by DCGAN that utilizes our proposed method.}
        \label{fig:gen_cifar10_a}
    \end{figure}
    
    % \begin{figure}[h]
    %     \centering
    %     \includegraphics[width=250px]{AAAI/generated_images/cifar10_b.png}
    %     \caption{CIFAR10-B}
    %     \label{fig:gen_cifar10_b}
    % \end{figure}
    
    \begin{figure}[h]
        \centering
        \captionsetup{width=.7\linewidth}
        \includegraphics[width=250px]{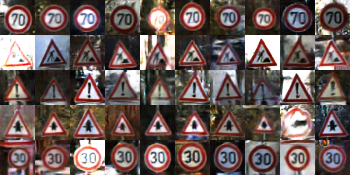}
        \caption{GTSRB images generated by DCGAN that utilizes our proposed method.}
        \label{fig:gen_gtsrb}
    \end{figure}
    
    \begin{figure}[ht!]
        \centering
        \captionsetup{width=.7\linewidth}
        \includegraphics[width=250px]{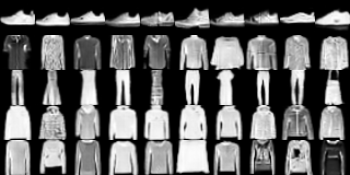}
        \caption{FashionMNIST images generated by DCGAN that utilizes our proposed method.}
        \label{fig:gen_fashion_mnist}
    \end{figure}
    
    \begin{figure}[h]
        \centering
        \captionsetup{width=.7\linewidth}
        \includegraphics[width=250px]{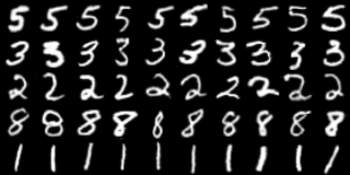}
        \caption{MNIST images generated by DCGAN that utilizes our proposed method.}
        \label{fig:gen_mnist}
    \end{figure}

    \clearpage

\end{document}